\documentclass[lettersize,journal]{IEEEtran}
\usepackage{amsmath,amsfonts}
\usepackage{algorithmic}
\usepackage{array}
\usepackage[caption=false,font=normalsize,labelfont=sf,textfont=sf]{subfig}
\usepackage{textcomp}
\usepackage[colorlinks,
            linkcolor=red,
            anchorcolor=blue,
            citecolor=green
            ]{hyperref}
\usepackage{stfloats}
\usepackage{url}
\usepackage{color}
\usepackage{verbatim}
\usepackage{graphicx}
\usepackage{overpic}
\usepackage{booktabs}
\usepackage{multirow}
\usepackage{bbding}
\usepackage[table]{xcolor}
\definecolor{seagreen}{RGB}{46,139,87}
\definecolor{MidLabel}{RGB}{255,231,198}       
\definecolor{OursLabel}{RGB}{205,244,205}      
\definecolor{OursLabel2}{RGB}{255,205,205}      
\definecolor{AltLabel}{RGB}{210,230,255}       

\newcommand{\labA}[2]{\colorbox{#1}{\hspace{3pt}\scriptsize #2\hspace{3pt}}}

\usepackage{tikz}       
\usetikzlibrary{spy,calc,positioning}
\newcommand{\ZoomCellSpyCircle}[8]{%
\begin{minipage}[t]{#2}\centering
\begin{tikzpicture}[
spy using outlines={circle,
  magnification=1.8,
  size=#8,
  connect spies,
  every spy on node/.append style={
    draw=#4,
    line width=0.6pt,
    shape=circle,
    minimum size=#7,
    inner sep=0pt
  },
  every spy in node/.append style={
    draw=#4,
    line width=0.6pt,
    shape=circle,
    inner sep=0pt,
    path picture={
      \pgfmathsetlengthmacro{\InsetR}{0.5*#8}%
      \clip (path picture bounding box.center) circle[radius=\InsetR];
    }
  },
  spy connection path={
    \draw[#4, line width=0.6pt] (tikzspyonnode) -- (tikzspyinnode);
  }
}
]
  \node[anchor=south west, inner sep=0, outer sep=0] (img) at (0,0)
    {\includegraphics[width=\linewidth]{#1}};

  \path (img.south west) -- (img.south east) coordinate[pos=#5] (Xb);
  \path (img.north west) -- (img.north east) coordinate[pos=#5] (Xt);
  \path (Xb) -- (Xt) coordinate[pos=#6] (CROP);

  \spy on (CROP) in node[anchor=north east] at ($(img.north east)+(-0.2,-0.2)$);
\end{tikzpicture}

\vspace{-4pt}
#3
\end{minipage}%
}

\def\BibTeX{{\rm B\kern-.05em{\sc i\kern-.025em b}\kern-.08em
    T\kern-.1667em\lower.7ex\hbox{E}\kern-.125emX}}
\usepackage{balance}

\begin{document}
\title{RSGMamba: Reliability-Aware Self-Gated State Space Model for Multimodal Semantic Segmentation}

\author{
Guoan Xu, 
Yang Xiao, 
Guangwei Gao,~\IEEEmembership{Senior Member,~IEEE},
Dongchen Zhu,~\IEEEmembership{Member,~IEEE},
Guo-Jun Qi,~\IEEEmembership{Fellow,~IEEE},
and Wenjing Jia,~\IEEEmembership{Member,~IEEE}

\thanks{Guoan Xu, Yang Xiao, and Wenjing Jia are with the Faculty of Engineering and Information Technology, University of Technology Sydney, Sydney, NSW 2007, Australia (e-mail: xga\_njupt@163.com, 
yang.xiao-2@student.uts.edu.au, and Wenjing.Jia@uts.edu.au).}

\thanks{Guangwei Gao is with the PCA Lab, Key Laboratory of Intelligent Perception and Systems for High-Dimensional Information of Ministry of Education, School of Computer Science and Engineering, Nanjing University of Science and Technology, Nanjing 210094, China (e-mail: gwgao@njust.edu.cn).}

\thanks{Dongchen Zhu is with Bionic Vision Systems Laboratory, Shanghai Institute of Microsystem and Information Technology, Chinese Academy of Sciences, Shanghai 200050, China (e-mail: dchzhu@mail.sim.ac.cn).}

\thanks{Guo-Jun Qi is with the Research Center for Industries of the Future and the School of Engineering, Westlake University, Hangzhou 310024, China, and also with OPPO Research, Seattle, WA 98101 USA (e-mail: guojunq@gmail.com).}
}

\markboth{Journal of \LaTeX\ Class Files,~Vol.~18, No.~9, September~2020}%
{How to Use the IEEEtran \LaTeX \ Templates}

\maketitle

\begin{abstract}
Multimodal semantic segmentation has emerged as a powerful paradigm for enhancing scene understanding by leveraging complementary information from multiple sensing modalities (e.g., RGB, depth, and thermal). However, existing cross-modal fusion methods often implicitly assume that all modalities are equally reliable, which can lead to feature degradation when auxiliary modalities are noisy, misaligned, or incomplete. In this paper, we revisit cross-modal fusion from the perspective of modality reliability and propose a novel framework termed the Reliability-aware Self-Gated State Space Model (RSGMamba). At the core of our method is the Reliability-aware Self-Gated Mamba Block (RSGMB), which explicitly models modality reliability and dynamically regulates cross-modal interactions through a self-gating mechanism. Unlike conventional fusion strategies that indiscriminately exchange information across modalities, RSGMB enables reliability-aware feature selection and enhancing informative feature aggregation. In addition, a lightweight Local Cross-Gated Modulation (LCGM) is incorporated to refine fine-grained spatial details, complementing the global modeling capability of RSGMB. Extensive experiments demonstrate that RSGMamba achieves state-of-the-art performance on both RGB-D and RGB-T semantic segmentation benchmarks, resulting 58.8\% / 54.0\% mIoU on NYUDepth V2 and SUN-RGBD (+0.4\% / +0.7\% over prior best), and 61.1\% / 88.9\% mIoU on MFNet and PST900 (up to +1.6\%), with only 48.6M parameters, thereby validating the effectiveness and superiority of the proposed approach.


\end{abstract}

\begin{IEEEkeywords}
Segmentation, Reliability-Aware Fusion, State Space Model (SSM), Global–Local Feature Enhancement
\end{IEEEkeywords}

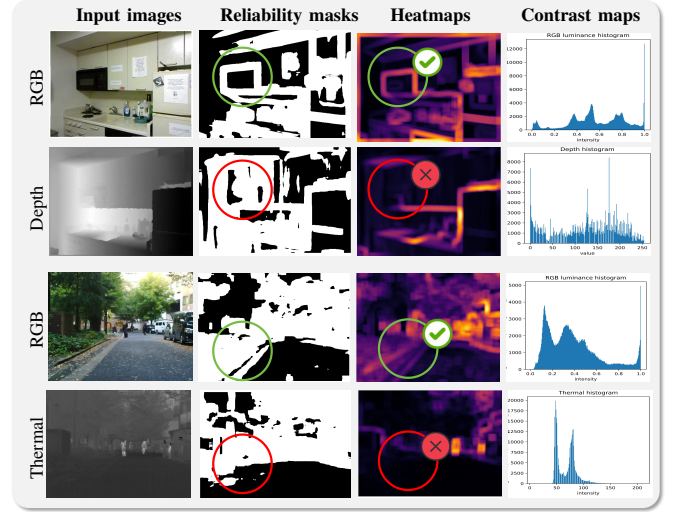
\begin{figure}[t]
   \centering
       \centering
       \begin{overpic}[width=0.48\textwidth]{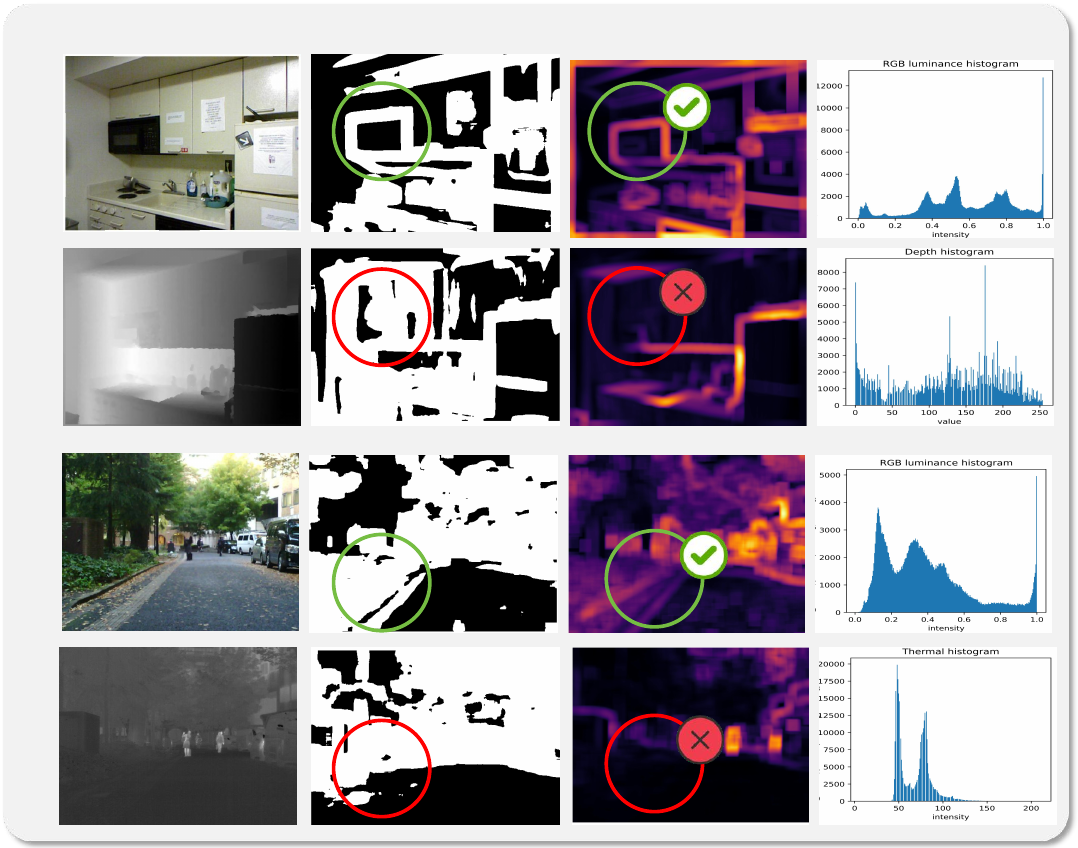}
        \put(10,75){\scriptsize \textbf{Input images}}
        \put(32,75){\scriptsize \textbf{Reliability masks}}
        \put(58,75){\scriptsize \textbf{Heatmaps}}
        \put(78,75){\scriptsize \textbf{Contrast maps}}
        \put(3,5){\rotatebox{90}{\scriptsize Thermal}}
        \put(3,25){\rotatebox{90}{\scriptsize RGB}}
        \put(3,43){\rotatebox{90}{\scriptsize Depth}}
        \put(3,62){\rotatebox{90}{\scriptsize RGB}}
    \end{overpic}
   \caption{For each RGB–X sample, we visualize the input image, the corresponding reliability mask, the cross-modal response heatmap, and the modality-wise value distribution. Green circles highlight regions where the modality provides reliable and consistent cues, while red circles denote unreliable or noisy regions. As shown, RGB luminance exhibits a smooth and continuous distribution, while depth and thermal signals present highly skewed and discrete histograms, with large portions of unreliable values. This pronounced statistical heterogeneity motivates the development of a more robust cross-modal fusion strategy that adaptively regulates cross-state readout based on modality uncertainty and cross-modal consistency.}
   \label{fig:heatmaps}
   \vspace{-2mm}
\end{figure}

\section{Introduction}
\label {sect:intro} 
Semantic segmentation aims to predict dense semantic labels at the pixel level and has become a fundamental task in the computer vision community. It plays a vital role in a wide range of applications, such as robotic manipulations~\cite{yuan2025roboengine, zhang2025zisvfm}, medical image analysis~\cite{xing2025diff,li2025u}, and autonomous driving~\cite {gao2022fbsnet}, all of which require accurate perception and understanding of the surrounding environment. Despite recent progress, existing models~\cite{yang2025golden,xu2024haformer,fu2025segman} remain vulnerable under adverse conditions, such as poor illumination and severe visual interference. To address these challenges, additional sensing modalities, including thermal and depth, are commonly incorporated. By exploiting complementary cues across modalities, the reliability and representational capacity of vision pipelines can be significantly improved.

Existing multimodal learning approaches~\cite{jia2024geminifusion, li2025stitchfusion, kim2024token,yin2025dformerv2,yin2025omnisegmentor} generally adopt two categories of fusion mechanisms: inner-interaction fusion (IIF) and outer-exchange fusion (OEF). These approaches can be further understood as three representative paradigms: simple addition, cross-attention-based interaction, and state-space-based exchange. IIF-based methods aim to enhance multimodal feature representations by explicitly modeling dependencies across different modalities. Early studies mainly relied on direct \textit{feature addition}~\cite{zhang2023delivering}, but they lack explicit cross-modal interaction, limiting their ability to exploit complementary information. More recent works increasingly incorporate attention mechanisms to facilitate cross-modal interactions. In particular, \textit{Cross-Attention} based designs, such as GeminiFusion~\cite{jia2024geminifusion} and TokenFusion~\cite{kim2024token}, have shown strong effectiveness in capturing fine-grained inter-modal relationships. However, modeling global dependencies through attention-driven interaction often incurs substantial computational and memory overhead.


DiffPixelFormer~\cite{gong2025diffpixelformer}, introduced a differential pixel-aware Transformer that applies Cross-Attention across modalities in a discrete manner, further reducing computational complexity. Nevertheless, such efficiency-oriented designs often come at the expense of fine-grained details and long-range contextual information, which may degrade segmentation performance, especially in scenarios that require precise global reasoning and subtle cross-modal cues. Moreover, interaction-based fusion methods often depend on large-scale RGB-X pretraining to fully exploit cross-modal interactions. A representative example is TUNI~\cite{guo2025tuni}, which first employs depth estimation models to generate pseudo-depth from ImageNet RGB images, thereby constructing paired RGB–depth training data, and then trains modality-specific backbones together with dedicated in-backbone fusion modules. While this paradigm can achieve strong performance, it significantly increases the demands on data preparation, training time, and computational resources, resulting in greater overall development complexity. Similar observations apply to DFormer~\cite{yin2024dformer} in the RGB-D setting, which attains competitive results but at the cost of substantially higher training overhead.

In contrast, outer-exchange-based fusion strategies can directly leverage pretrained RGB backbones, while the auxiliary modality branch often shares weights with the RGB stream, enabling plug-and-play fusion with significantly lower training cost. For example, ADBNet~\cite{xu2025adbnet} introduced an asymmetric encoder with enhanced depth processing and proposed CF-PVT (Conv-Former Pyramid Vision Transformer), which uses decomposed convolutional attention to strengthen depth feature extraction. Beyond CNN- and ViT-based designs, Sigma~\cite{wan2025sigma} explored state space modeling to avoid the quadratic complexity of self-attention and employs \textit{Cross-Mamba} to facilitate efficient interaction between RGB and depth features. Following a similar direction, CM-SSM~\cite{guo2025cross} extended the Cross-Mamba paradigm to RGB-T segmentation by further designing a scanning strategy that interleaves RGB and thermal tokens, thereby enabling more efficient sequence correlation modeling. Nevertheless, the efficiency of exchange-based fusion comes with more stringent fusion design requirements. Since auxiliary modalities are often noisy or unreliable, indiscriminate fusion may introduce feature contamination and degrade overall performance. As illustrated in Fig.~\ref{fig:heatmaps}, the reliability of auxiliary modalities varies significantly across spatial regions, highlighting the importance of reliability-aware cross-modal fusion.

Overall, existing multimodal fusion approaches still suffer from several inherent limitations.
(1) Interaction-based methods typically rely on large-scale RGB-X pretraining to enable effective cross-modal interaction, resulting in substantial training cost and data dependency.
(2) Exchange-based methods, while more efficient, often perform unconditional feature fusion and thus are highly sensitive to the quality of auxiliary modalities. Fundamentally, both paradigms overlook explicit modeling of modality reliability, making them vulnerable to noisy or misaligned auxiliary inputs. Therefore, the key challenge of multimodal fusion lies not only in how to exchange information across modalities, but also in how to selectively trust them.

To address these challenges, we propose a reliability-aware fusion framework, termed \textit{RSGMamba}, which explicitly models modality reliability during cross-modal interaction. Specifically, upon extracting multi-level features from the backbone, we introduce a reliability-aware fusion module to adaptively regulate cross-modal interactions between the RGB stream and the auxiliary modality. Unlike standard Cross-Mamba, which enforces unconditional cross-state coupling between modalities, our proposed Reliability-aware Self-Gated Mamba Block (RSGMB) explicitly controls cross-modal interactions at the feature level. In particular, it introduces: (i) an uncertainty-aware gate to estimate the reliability of each modality independently, and (ii) a consistency-aware gate to evaluate cross-modal agreement at each spatial location. These two gates jointly modulate the cross-state readout in the state space model, enabling adaptive control over the intensity of cross-modal information injection. Furthermore, we employ a low-rank cross projection with a learnable scaling factor, which constrains the fusion capacity and serves as an implicit regularizer. This design prevents overfitting to noisy auxiliary signals and encourages cross-modal fusion to function as a lightweight, reliability-driven calibration rather than a dominant transformation. In addition to global fusion, our designed Local Cross-Gated Modulation (LCGM) captures local cross-modal interactions to refine fine-grained spatial details and boundary structures. As a result, our approach effectively alleviates negative transfer caused by noisy or misaligned auxiliary modalities, leading to more robust RGB-X segmentation performance. Beyond long-sequence modeling, we further enhance RGB representations from a local perspective. By integrating global and local representations, the proposed method improves discrimination under complex backgrounds and noise. Our contributions can be summarized as follows:
\begin{itemize}
    \item We propose RSGMamba, a reliability-aware state space framework that explicitly models modality reliability and cross-modal consistency, enabling efficient global cross-modal interaction with linear complexity.
    \item We design a Reliability-aware Self-Gated Mamba Block (RSGMB) and a Local Cross-Gated Modulation (LCGM), 
    which jointly model global modality reliability and cross-modal consistency, while preserving fine-grained local structures.
    \item Extensive experiments on multiple RGB-D and RGB-T benchmarks, including NYUDepth v2, SUN RGB-D, MFNet, and PST900, demonstrate that our method achieves a competitive trade-off between accuracy and efficiency and consistently outperforms existing approaches.
\end{itemize}


\begin{figure*}
    \centering
    \includegraphics[width=1.0\linewidth]{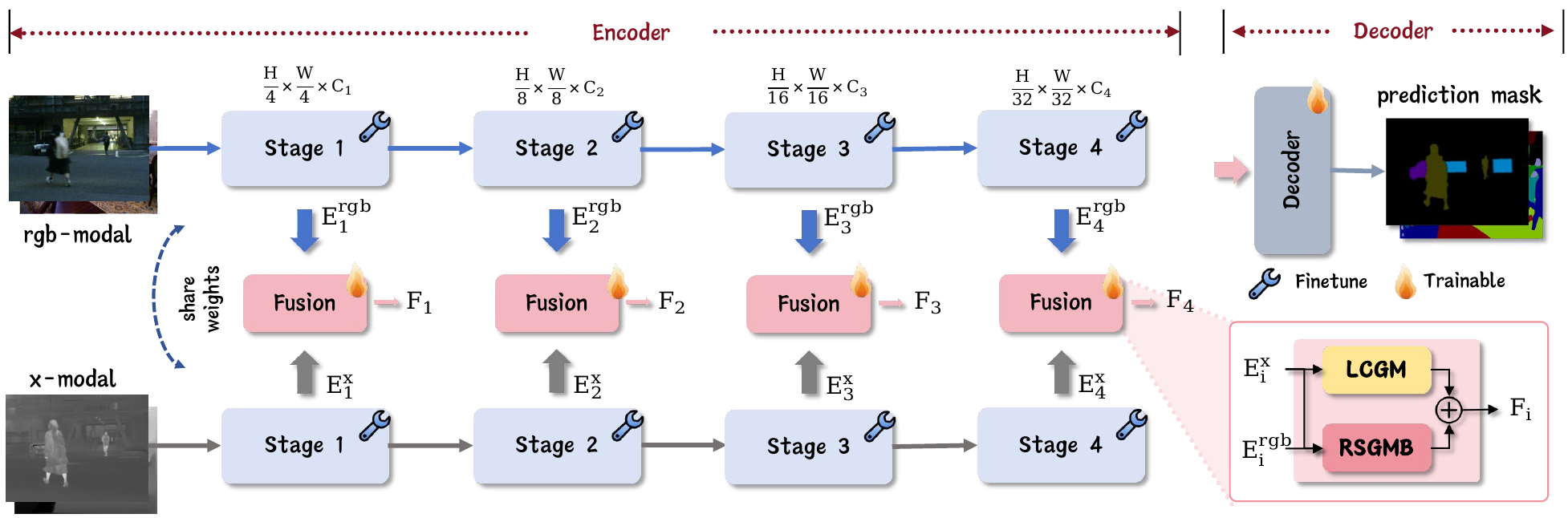}
    \caption{Overall architecture of our RSGMamba for RGB–X segmentation. RGB and auxiliary modalities are encoded through two weight-sharing encoders to produce multi-scale representations $E^{rgb}_i$ and $E^{x}_i$. At each stage, a fusion module combines cross-modal features using the proposed Local Cross-Gated Modulation (LCGM) and Reliability-aware Self-Gated Mamba Block (RSGMB), generating fused features $F_i$. Finally, the four-stage fusion features $F_1$, $F_2$, $F_3$, and $F_4$ are fed into the decoder module for final segmentation mask generation.}
    \label{fig:model}
\end{figure*}

\section{Related Works}
\label {sect:relatedwork}

\subsection{Multimodal Semantic Segmentation}
Multimodal semantic segmentation~\cite{zhang2023cmx,zhang2023delivering,yin2025dformerv2,bai2025dcanet} extends conventional RGB-based perception by incorporating complementary modalities such as thermal, depth, and LiDAR, which provide informative cues beyond appearance alone. Effectively leveraging these heterogeneous signals requires carefully designed feature extractors and fusion mechanisms, making cross-modal representation learning a central research focus in this area. 

In RGB-Depth semantic segmentation, effectively integrating RGB and depth representations has been shown to substantially improve semantic segmentation performance. One line of research follows the inner-interaction fusion paradigm, which integrates multimodal representations through explicit cross-modal interactions, such as Cross-Attention mechanisms or feature concatenation. These methods typically perform pretraining on large-scale ImageNet-1K RGB–D paired data and then fine-tune the model on downstream RGB–D benchmarks using a unified network architecture~\cite{yin2025dformerv2,yin2025omnisegmentor,yin2026dformer++}. However, their performance heavily depends on the quality and scale of the pretraining datasets, and the two-stage training pipeline also incurs substantial computational and training costs. An alternative line of work explores outer-exchange fusion, in which modality-specific features are encoded separately and later integrated in the decoder through dynamic fusion mechanisms. Methods including CMX~\cite{zhang2023cmx}, TokenFusion~\cite{kim2024token}, GeminiFusion~\cite{jia2024geminifusion}, and Sigma~\cite{wan2025sigma} have demonstrated strong performance on RGB–D semantic segmentation benchmarks. \textit{Nevertheless, these methods typically assume equal reliability across modalities. Since depth data is often affected by noise and invalid measurements, naive or symmetric fusion may introduce cross-modal interference, thereby limiting potential performance gains.}

In RGB-Thermal semantic segmentation, early dual-stream frameworks such as MFNet~\cite{ha2017mfnet} and RTFNet~\cite{sun2019rtfnet} demonstrated the feasibility of multimodal fusion, although their simplistic element-wise integration strategies did not adequately address cross-modal inconsistencies. Zhou et al.~\cite{zhou2021gmnet} designed an adaptive cross-modal fusion framework that applies spatial attention to low-level representations while leveraging channel-wise attention at higher semantic stages. By using multimodal features to construct query–key–value triplets, MFTNet~\cite{zhou2022multispectral} and MCNet-T~\cite{jiang2024mirror} effectively promoted feature-level interactions across modalities. DFormerv2~\cite{yin2025dformerv2} exploited geometric attention to extract informative cues from the thermal modality and inject them into the network for enhanced representation learning. Inspired by this design philosophy, TUNI~\cite{guo2025tuni} adopted a similar formulation but replaced real thermal inputs with RGB-derived pseudo-modality representations, enabling geometry-aware feature enhancement without relying on additional sensor data. To reduce the computational overhead of Cross-Attention, Sigma~\cite{wan2025sigma} and CMSSM~\cite{guo2025cross} introduced Mamba-based architectures into RGB–T semantic segmentation and adopted an outer-exchange fusion paradigm for cross-modal interaction. \textit{Although these approaches improve efficiency, they generally overlook modality reliability by treating RGB and thermal features symmetrically during scanning and information exchange, which may result in suboptimal fusion when one modality is degraded or noisy.}

\subsection{State Space Models}
Recent advances in computer vision have increasingly adopted state space models (SSMs) to address long-range dependency modeling. Among them, the structured state space sequence model~\cite{gu2021efficiently} is particularly notable for enabling effective long-range representation learning with linear computational complexity, making it a compelling alternative to attention-based approaches. These favorable properties have made state space models increasingly attractive for vision tasks. By enabling efficient global dependency modeling through multi-directional scanning, recent works such as Vision Mamba~\cite{gu2024mamba} and VMamba~\cite{liu2024vmamba} have demonstrated strong effectiveness. Meanwhile, SSMs~\cite{zhu2024vision} have been increasingly explored in dense prediction tasks, including segmentation, restoration, and super-resolution, highlighting their versatility. Nevertheless, recent studies often adopt state space models as generic building blocks without fully tailoring their design to the specific requirements of individual vision tasks. In addition, the application of SSMs to multimodal settings has received relatively limited attention. Recent studies that introduce Mamba~\cite{gu2021efficiently} into RGB-based multi-modal semantic segmentation mainly include Sigma~\cite{wan2025sigma} and CMSSM~\cite{guo2025cross}. These methods fuse multimodal features using a Cross-Mamba interaction strategy and have demonstrated notable effectiveness. However, their fusion mechanisms still have room for improvement, as they do not explicitly account for the reliability of the exchanged information. In practice, features from different modalities are often fused in a blind and symmetric manner, which may introduce cross-modal interference when one modality is noisy or unreliable. \textit{To address this limitation, we explicitly incorporate modality reliability into the fusion process. Specifically, we introduce a self-gating mechanism that first evaluates the reliability of each modality independently and then assesses cross-modal consistency. This reliability-aware and consistency-guided fusion strategy enables more selective cross-modal interaction, leading to improved segmentation performance.}

\begin{figure*}
    \centering
    \includegraphics[width=1.0\linewidth]{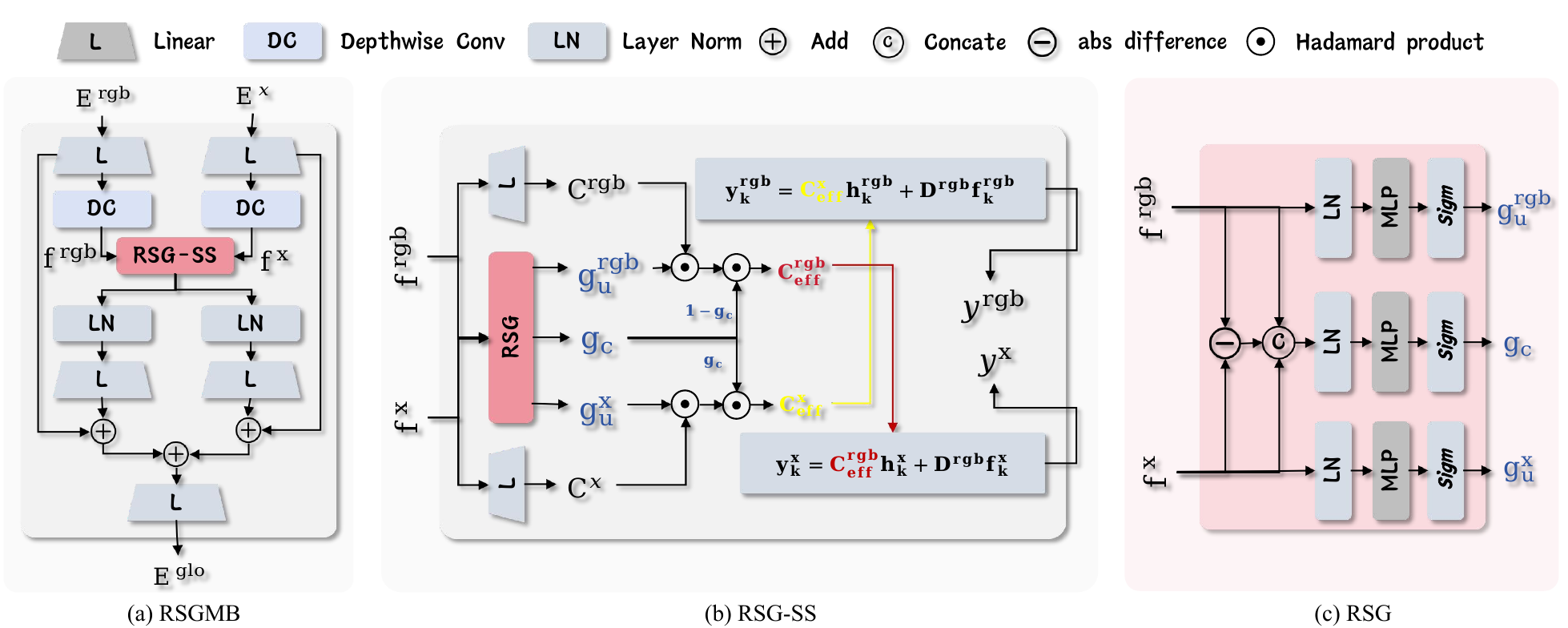}
    \caption{(a) Overall architecture of the proposed RSGMB, where RGB and auxiliary-modal features are processed in parallel and fused via a reliability-controlled cross-state mechanism. (b) Illustration of the proposed RSG-SS module. For simplicity, certain implementation details are omitted, while the key computational components that distinguish our method from prior designs are highlighted. (c) Internal design of the reliability self-gating module (RSG), which estimates modality-wise uncertainty and cross-modal consistency to dynamically regulate cross-state readout in the Mamba-based state-space model.}
    \label{fig:rsgmb}
\end{figure*}

\begin{figure}
    \centering
    \includegraphics[width=1.0\linewidth]{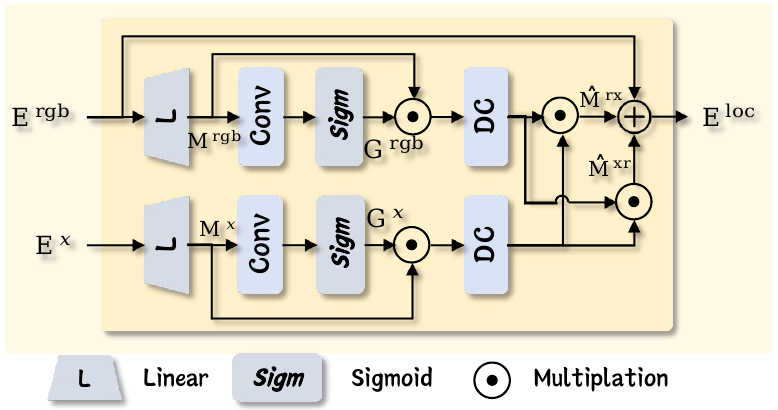}
    \caption{Structure of the Local Cross-Gated Modulation (LCGM). The RGB and auxiliary features are first aligned through linear projection and cross-modal gating. The resulting features are then processed with depthwise convolutions to capture local spatial structures, yielding the enhanced feature representation $E^{loc}$.}
    \label{fig:lem}
\end{figure}

\section{Methodology}
\label {sect:method}
\subsection{Preliminary}

Structured state space sequence models (S4) are inspired by the continuous system, which maps a 1-D sequence $x(t)\in \mathbb{R}$ into $y(t)\in \mathbb{R}$, through a hidden state $h(t)\in \mathbb{R^{N}}$. The continuous system can be summarized as follows: 
\begin{equation}
    h'(t) = Ah(t)+Bx(t),
    y(t) = Ch(t)+Dx(t),
\end{equation}
where $A\in\mathbb{R}^{N\times N}$, $B\in\mathbb{R}^{N\times 1}$, $C\in\mathbb{R}^{1 \times N}$, and $D\in\mathbb{R}$ are the system parameters. To accommodate discrete inputs including visual and textual sequences, state space models usually rely on Zero-Order Hold (ZOH) discretization to map the input sequence ${x_1,x_2,...,x_K}$ into the output sequence ${y_1,y_2,...,y_K}$. Specifically, a timescale parameter $\Delta$
to transform the continuous parameters $A$, $B$ to its corresponding discrete versions $\overline{A}$, $\overline{B}$, this process can be formulated as
\begin{equation}
    \begin{split}
    &\overline{A} = exp(\Delta A), \overline{C} = C, \overline{D} = D,\\
    &\overline{B} = (\Delta A)^{-1}(exp(A)-I)\Delta B.
    \end{split}
\end{equation}
The discretized version can be rewritten as
\begin{equation}
    h_k = \overline{A}h_{k-1}+Bx_k,
    y_k = \overline{C}h_k+\overline{D}x_k.
\end{equation}
Selective State Space Models (S6), commonly known as Mamba, introduce input-conditioned parameterization that dynamically modulates the state space operators, resulting in a more expressive formulation for sequential data. Specifically, the state transition and projection matrices $B\in \mathbb{R}^{B \times L \times N}$, $C\in \mathbb{R}^{B \times L \times N}$, and $\Delta\in \mathbb{R}^{B \times L \times D}$ are directly generated from the input sequence $x\in \mathbb{R}^{B \times L \times D}$. By leveraging selective state space modeling, Mamba achieves linear scalability with respect to sequence length while exhibiting strong capability in vision modeling tasks.


\subsection{Overall Architecture}
As illustrated in Fig.~\ref{fig:model}, our framework consists of a pretrained encoder, a decoder, and a core fusion module for robust cross-modal interaction. We adopt SegMAN~\cite{fu2025segman} as the backbone due to its favorable trade-off between lightweight design and strong performance.

The RGB encoder and the auxiliary-modal encoder share the same network parameters, ensuring consistent representation learning across modalities. Given an input RGB image $\mathbf{I}^{rgb} \in \mathbb{R}^{H\times W \times 3}$ and its corresponding auxiliary modality $\mathbf{I}^{x} \in \mathbb{R}^{H\times W \times 3}$ (e.g., depth or thermal), the shared encoder produces four hierarchical feature levels for each modality, denoted as $\mathbf{E}_i^{rgb} \in \mathbb{R}^{\frac{H}{2^{i+1}} \times \frac{W}{2^{i+1}} \times C_i}$ and $\mathbf{E}_i^{x} \in \mathbb{R}^{\frac{H}{2^{i+1}} \times \frac{W}{2^{i+1}} \times C_i}$.

At each corresponding feature level, we perform cross-modal fusion by decomposing the process into global fusion and local fusion, which capture complementary contextual cues with different receptive fields. Specifically, the global fusion branch employs the proposed Reliability-aware Self-Gated Mamba Block (RSGMB) to model long-range dependencies and regulate cross-modal interactions in a reliability-aware manner at the state-space level. In parallel, the local fusion branch utilizes a Local Cross-Gated Modulation (LCGM) to emphasize fine-grained spatial details and local structural information.

The outputs from the global and local fusion branches are then combined via element-wise addition, enabling complementary integration of global contextual reasoning and local detail refinement. The fused multi-level features are finally forwarded to the decoder to generate the final segmentation prediction. This dual-branch fusion strategy allows our model to jointly exploit global cross-modal dependencies and local spatial cues, leading to more robust and accurate segmentation under challenging cross-modal conditions.

\subsection{Fusion Modules}
Intuitively, modality reliability reflects the confidence of each modality at different spatial locations. In real-world scenarios, auxiliary modalities (e.g., depth or thermal) often contain noise, missing values, or misalignment. Therefore, effective fusion requires suppressing unreliable responses while preserving informative regions.
\textbf{Reliability-aware Self-Gated Mamba Block (RSGMB).} As illustrated in Fig.~\ref{fig:rsgmb}, we propose a Reliability-Aware Self-Gated Mamba Block (RSGMB) to enable robust and adaptive cross-modal fusion in RGB-X segmentation tasks. The core limitation of existing fusion modules is that they either indiscriminately inject auxiliary cues or rely on dense interaction, making them vulnerable to noisy and misaligned modalities, while our RSGMB performs cross-modal interaction within a Mamba-based state-space model (SSM) and introduces an explicit reliability-driven self-gating mechanism to dynamically regulate the strength of cross-modal information flow. Specifically, given RGB and auxiliary features extracted at the same semantic level, $\textbf{E}^{\text{rgb}}, \textbf{E}^{\text{x}} \in \mathbb{R}^{H \times W \times C}$, RSGMB processes both modalities in parallel and performs cross-state fusion at the readout stage rather than directly mixing feature representations. 

Cross-modal modeling often relies on high-dimensional linear projections, which can substantially increase model parameters and computational cost. To address this issue, we adopt low-rank linear projections for both modality-specific and cross-modal transformations. A standard linear mapping $\mathbf{W} \in \mathbb{R}^{d_{\text{out}} \times d_{\text{in}}}$ is approximated by a low-rank decomposition:
\begin{equation}
    \mathbf{W} \approx \alpha \, \mathbf{W}_{\text{up}} \mathbf{W}_{\text{down}}, 
\end{equation} 
\begin{equation}
    \mathbf{W}_{\text{down}}\in \mathbb{R}^{r \times d_{\text{in}}},
    \mathbf{W}_{\text{up}}\in \mathbb{R}^{d_{\text{out}} \times r},
\end{equation}
where $r \ll \min(d_{\text{in}}, d_{\text{out}})$ is the low-rank dimension, and $\alpha$ is a learnable scaling factor. This design reduces the parameter complexity from
$\mathcal{O}(d_{\text{in}}d_{\text{out}})$ to $\mathcal{O}(r(d_{\text{in}}+d_{\text{out}}))$ while preserving expressive capacity. Moreover, the low-rank formulation improves training stability when applied to cross-modal fusion. 

After projection, the RGB and auxiliary features are modeled using parallel Mamba-based state-space models. For the $k$-th token in the flattened spatial sequence, the state updates are defined as
\begin{equation}
    \begin{split}
    &\mathbf{h}_k^{\text{rgb}} = \mathbf{A}^{\text{rgb}}\mathbf{h}_{k-1}^{\text{rgb}}+\mathbf{B}^{\text{rgb}}\mathbf{f}_{k}^{\text{rgb}}, \\
    &\mathbf{h}_k^{\text{x}} = \mathbf{A}^{\text{x}}\mathbf{h}_{k-1}^{\text{x}}+\mathbf{B}^{\text{x}}\mathbf{f}_{k}^{\text{x}},
    \end{split}
\end{equation}
where $\textbf{A}$ and $\textbf{B}$ are parameterized following the Mamba formulation and efficiently computed via selective scan. Unlike standard Mamba, RSGMB does not employ a fixed output projection. Instead, the output readout is dynamically modulated by reliability-aware gates. To assess the reliability of each modality independently, we introduce \textbf{modality-wise uncertainty-aware gates:}
\begin{equation}
    \mathbf{g}_{\text{u}}^{\text{rgb}} = \sigma\!\left( \mathcal{G}_{\text{u}}^{\text{rgb}}(\mathbf{f}^{\text{rgb}}) \right), 
    \mathbf{g}_{\text{u}}^{x} = \sigma\!\left( \mathcal{G}_{\text{u}}^{x}(\mathbf{f}^{x}) \right),
\end{equation}
where $\mathcal{G}_{\text{u}}$ consists of Layer Normalization followed by a lightweight MLP. The uncertainty-aware gates estimate modality-wise reliability by measuring the stability of feature responses. Regions with high uncertainty (e.g., noisy or missing values) tend to produce lower gating scores, thereby reducing their influence during cross-modal interaction.

Relying solely on single-modality reliability is insufficient to determine whether cross-modal fusion should be performed. We therefore further introduce a \textbf{cross-modal consistency-aware gate:}
\begin{equation}
    \mathbf{g}_{\text{c}} = \sigma\!\left( \mathcal{G}_{\text{c}}([\mathbf{f}^{\text{rgb}},\mathbf{f}^{\text{x}},|\mathbf{f}^{\text{rgb}}-\mathbf{f}^{\text{x}}|]) \right), 
\end{equation}
where $\mathcal{G}_{\text{c}}$ consists of Layer Normalization followed by a lightweight MLP. It explicitly models inter-modal agreement by incorporating both raw features and their absolute difference. The consistency-aware gate captures cross-modal agreement. When features from different modalities are consistent, the gating value is increased, promoting reliable information exchange. Conversely, inconsistent responses are suppressed to avoid feature contamination. Based on the estimated reliability and consistency, we construct dynamic cross-state matrices:
\begin{equation}
\begin{aligned}
\mathbf{C}_{\text{eff}}^{\text{rgb}} 
&= \mathbf{g}_{\text{u}}^{\text{rgb}} \cdot \left( 1 - \mathbf{g}_{\text{c}} \right) \cdot \mathbf{C}_{\text{rgb}}, \\
\mathbf{C}_{\text{eff}}^{x} 
&= \mathbf{g}_{\text{u}}^{x} \cdot \mathbf{g}_{\text{c}} \cdot \mathbf{C}_{x}.
\end{aligned}
\end{equation}

These effective matrices are then used in the selective scan to produce the final outputs:
\begin{equation}
\begin{aligned}
\mathbf{y}_{\text{k}}^{\text{rgb}} 
&= \mathbf{C}_{\text{eff}}^{\text{rgb}}\mathbf{h}_{\text{k}}^{\text{rgb}} + \mathbf{D}^{\text{rgb}}\mathbf{f}_{\text{k}}^{\text{rgb}}, \\
\mathbf{y}_{\text{k}}^{\text{x}} 
&= \mathbf{C}_{\text{eff}}^{\text{x}}\mathbf{h}_{\text{k}}^{\text{x}} + \mathbf{D}^{\text{x}}\mathbf{f}_{\text{k}}^{\text{x}}.
\end{aligned}
\end{equation}

This formulation allows RSGMB to adaptively regulate cross-modal fusion based on modality reliability and cross-modal consistency. When the auxiliary modality is unreliable, RSGMB naturally falls back to single-modality modeling, whereas consistent modalities trigger stronger cross-modal interactions. This design enables token-wise, non-uniform fusion without requiring any additional supervision.

Compared with Cross-Attention and fixed Cross-Mamba fusion, RSGMB offers three key advantages: (i) dynamic cross-state readout instead of static fusion weights, (ii) explicit modeling of modality reliability and cross-modal consistency, and (iii) a parameter-efficient design enabled by low-rank projections. These properties make RSGMB particularly well-suited for real-world RGB–X segmentation, where auxiliary modalities are often imperfect. This design enables dynamic modulation of cross-modal information flow, where the contribution of each modality is quantitatively controlled by the learned gating values.

\textbf{Local Cross-Gated Modulation (LCGM).}
Global fusion mechanisms are effective at modeling long-range dependencies, but they often fail to preserve fine-grained local structures that are essential for precise boundary delineation. To complement global fusion, we introduce a Local Cross-Gated Modulation (LCGM), which enables spatially adaptive, token-wise modulation between modalities at each resolution level. Given the RGB feature $\textbf{E}^{\text{rgb}}$ and auxiliary-modal feature $\textbf{E}^{\text{x}}$ at the same resolution, both features are first projected into a reduced embedding space via linear layers: 
\begin{equation}
    \mathbf{M}^{\text{rgb}}=\text{W}^{\text{rgb}}\text{E}^{\text{rgb}},
    \mathbf{M}^{\text{x}}=\text{W}^{\text{x}}\text{E}^{\text{x}},
\end{equation}
where $\text{W}^{\text{rgb}}$ and $\text{W}^{\text{x}}$ denote the linear projection matrices.
To adaptively modulate local responses within each modality, we generate modality-specific gating maps using shallow convolutional layers followed by sigmoid activation: 
\begin{equation}
    \mathbf{G}^{\text{rgb}}=\sigma(\text{Conv}(\mathbf{M}^{\text{rgb}})),
    \mathbf{G}^{\text{x}}=\sigma(\text{Conv}(\mathbf{M}^{\text{x}})),
\end{equation}
where $\sigma$ represents the Sigmoid function.
The gate is then applied to its own feature via element-wise multiplication, enabling self-adaptive local filtering:  
\begin{equation}
    \mathbf{\hat{M}}^{\text{rgb}}=\mathbf{DConv}(\mathbf{M}^{\text{rgb}}\odot \mathbf{G}^{\text{rgb}}),
    \mathbf{\hat{M}}^{\text{x}}=\mathbf{DConv}(\mathbf{M}^{\text{x}}\odot \mathbf{G}^{\text{x}}),
\end{equation}
where $\odot$ denotes element-wise multiplication, and $DConv$ means Depth-wise Convolution. The self-gated features are subsequently refined by depthwise convolutions to efficiently capture local spatial patterns while maintaining low computational overhead. After this local refinement stage, cross-modal interaction is introduced through element-wise multiplication between the two modality branches: 
\begin{equation}
    \mathbf{\hat{M}}^{\text{rx}}=\mathbf{\hat{M}}^{\text{rgb}} \odot \mathbf{\hat{M}}^{\text{x}},
    \mathbf{\hat{M}}^{\text{xr}}=\mathbf{\hat{M}}^{\text{x}} \odot \mathbf{\hat{M}}^{\text{rgb}}.
\end{equation}
Finally, the fused local response is added to the RGB branch via a residual connection, producing the enhanced local representation $\textbf{E}^{\text{loc}}$. 
\begin{equation}
    \textbf{E}^{\text{loc}}=\mathbf{E}^{\textbf{rgb}} + \mathbf{\hat{M}}^{\text{rx}} + \mathbf{\hat{M}}^{\text{xr}}.
\end{equation}

This design allows LCGM to selectively emphasize locally consistent structures while effectively suppressing unreliable auxiliary cues, thereby providing robust fine-grained enhancement that complements global fusion.


\begin{table*}[!t]
\begin{center}
\caption{Qualitative comparisons on NYUDepth V2 and SUN-RGBD datasets. All the backbones are pre-trained on ImageNet-1K. The \colorbox[RGB]{255,205,205}{best} and \colorbox[RGB]{255,231,198}{second} best results for each metric are color-coded.}
\label{RGBD}
\begin{tabular}{lccccccccc}
\toprule
\multirow{2}{*}{\textbf{Model}} & \multirow{2}{*}{\textbf{Publication}}  & \multirow{2}{*}{\textbf{Backbone}} & \multirow{2}{*}{\textbf{Params.}$\downarrow$} & \multicolumn{3}{c}{\textbf{NYUDepthv2}} & \multicolumn{3}{c}{\textbf{SUN-RGBD}}\\ \cline{5-7} \cline{8-10} 
&&& & \textbf{Input size} & \textbf{Flops}$\downarrow$ & \textbf{mIoU}$\uparrow$ & \textbf{Input size} & \textbf{Flops}$\downarrow$ & \textbf{mIoU}$\uparrow$ \\ 
\midrule 

TokenFusion~\cite{wang2022multimodal} & CVPR'22  & MiT-B3 & 45.9M & $480 \times 640$ & 94.4G & 54.2 & $530 \times 730$ & \cellcolor[RGB]{255,231, 198}122.1G & 51.0 \\
CMX~\cite{zhang2023cmx} & TITS'23  & MiT-B4 &  139.9M & $480 \times 640$ & 134.3G & 56.3 & $530 \times 730$ & 173.8G & 52.1\\
DFNet~\cite{yang2025difference} & TII'24 & MiT-B2 & \cellcolor[RGB]{255,231, 198}41.2M & $480 \times 640$ & \cellcolor[RGB]{255,205,205}35.1G & 56.1 & $530 \times 730$ & - & 51.6\\
DCANet~\cite{bai2025dcanet} & PR'25 & VMamba & 123.8M & $480 \times 640$ & 138.5G & 53.3 & $530 \times 730$ & - & 49.6 \\
ECMRN~\cite{jia2025ecmrn} & KBS'25 & DFormer & 68.6M & $480 \times 640$ & 135.3G &54.6 & $530 \times 730$ & 178.5G & 50.5 \\
Sigma~\cite{wan2025sigma} & WACV'25  &VMamba-T & 48.3M & $480 \times 640$ & 90.4G & 53.9 & $480 \times 640$ & 90.4G & 50.0 \\
ADBNet~\cite{xu2025adbnet} & KBS'25 &ConvNext-tiny  & 45.9M & $480 \times 640$ &\cellcolor[RGB]{255,231, 198}52.7G & 56.0 & $530 \times 730$ & - & 49.6 \\
DiffPixelFormer~\cite{gong2025diffpixelformer} & TMM'25 & MiT-B3 & 85.4M & $480 \times 640$ & 154.8G & 56.3 & $530 \times 730 $ & 205.7G & 52.8 \\
\midrule
\textbf{RSGMamba-S} & \textbf{UR'26}  & \textbf{SegMAN-S} & \cellcolor[RGB]{255,205,205}\textbf{28.1M} & $480 \times 640$ & \textbf{58.5G} & \textbf{56.4} & $ 530 \times 730$ & \cellcolor[RGB]{255,205,205}\textbf{75.8G} & \textbf{52.9} \\
\midrule
CMX~\cite{zhang2023cmx} & TITS'23  & MiT-B5 &  181.1M & $480 \times 640$ & 167.8G & 56.9 & $530 \times 730$ & 217.6G & 52.4\\
CMNext~\cite{zhang2023delivering} & CVPR'23  & MiT-B4 & 119.6M & $480 \times 640$ & 131.9G & 56.9 & $530 \times 730$ & 170.3G & 51.9\\
GeminiFusion~\cite{jia2024geminifusion} & ICML'24 & MiT-B3 & 75.8M & $480 \times 640$ & 174.0G & 56.8 & $530 \times 730$ & - & 52.7 \\
PrimKD~\cite{hao2024primkd} & MM'24 & MiT-B4 & 139.9M & $480 \times 640$ &134.3G &57.8 & $530 \times 730$ & 173.8G & 52.5\\
DFNet~\cite{yang2025difference} & TII'24 & MiT-B4 & 108.8M & $480 \times 640$ & 101.0G & 57.9 & $530 \times 730$ & - & 52.4\\
Sigma~\cite{wan2025sigma} & WACV'25  &VMamba-S & 69.8M & $480 \times 640$ & 139.8G & 57.0 & $480 \times 640$ & 139.8G & 52.4 \\
ADBNet~\cite{xu2025adbnet} & KBS'25 &ConvNext-base  & 299.1M & $480 \times 640$ &361.7G & 57.6 & $530 \times 730$ & - & 52.7 \\
DFormerV2~\cite{yin2025dformerv2} & CVPR'25  & DFormerV2-L & 95.5M & $480 \times 640$ & 124.1G & \cellcolor[RGB]{255,231,198}58.4 & $530 \times 730$ & 160.5G & \cellcolor[RGB]{255,231, 198}\textbf{53.3}\\
\midrule
\textbf{RSGMamba-B} & \textbf{UR'26}  & \textbf{SegMAN-B} & \textbf{48.6M} & $480 \times 640$ & \textbf{130.2G} & \cellcolor[RGB]{255,205,205}\textbf{58.8} & $ 530 \times 730$ & \textbf{168.4G} & \cellcolor[RGB]{255,205,205}\textbf{54.0} \\

\bottomrule
\end{tabular}
\end{center}
\end{table*}

\begin{figure*}[!t]
\centering
\setlength{\tabcolsep}{0pt}
\renewcommand{\arraystretch}{1.0}

\begin{tabular}{@{}cccccc@{}}
\ZoomCellSpyCircle{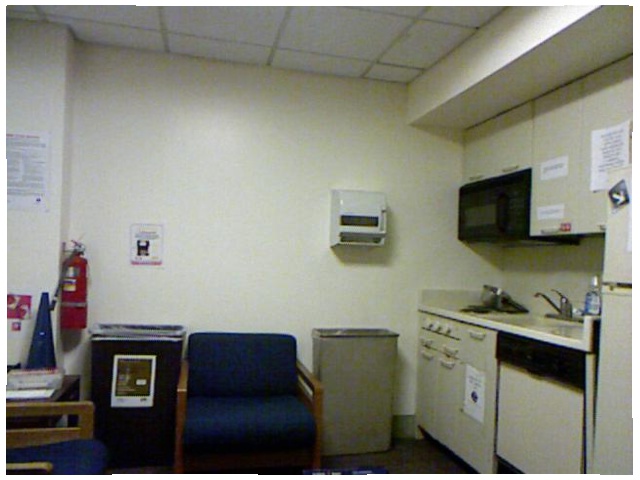}{0.162\textwidth}
  {\labA{MidLabel} {RGB img}}
  {cyan!60!black}{0.38}{0.18}  {1cm}{1.5cm}
&
\ZoomCellSpyCircle{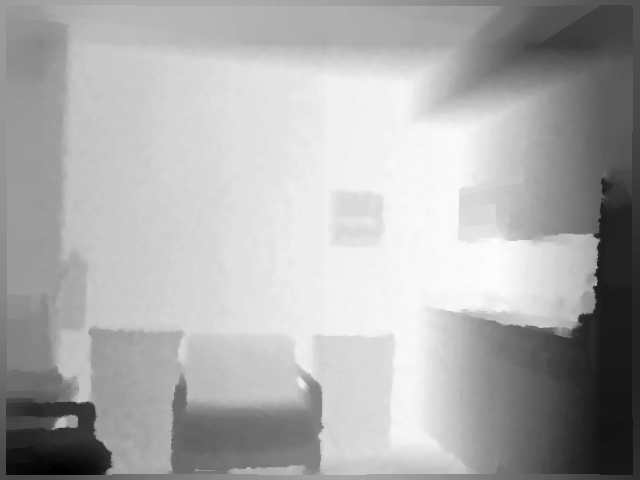}{0.162\textwidth}
  {\labA{MidLabel}{Depth img}}
  {orange}{0.38}{0.18}  {1cm}{1.5cm}
&
\ZoomCellSpyCircle{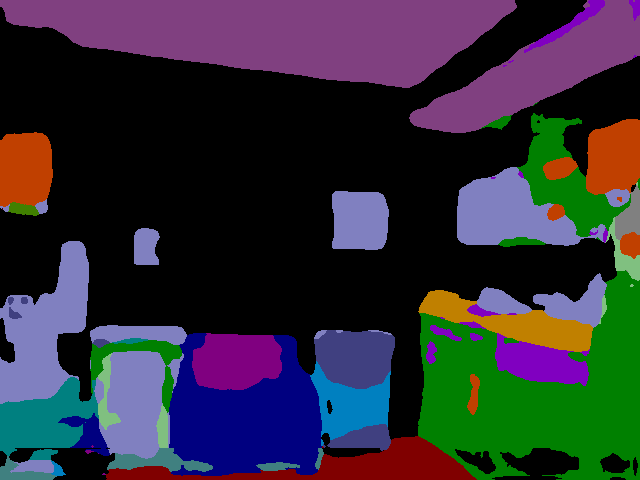}{0.162\textwidth}
  {\labA{MidLabel}{Sigma~\cite{wan2025sigma} (53.9\%)}}
  {orange}{0.38}{0.18}  {1cm}{1.5cm}
&
\ZoomCellSpyCircle{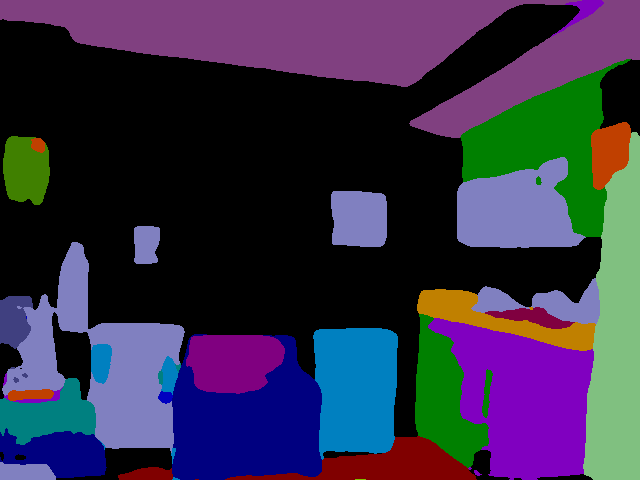}{0.162\textwidth}
  {\labA{MidLabel}{ECMRN~\cite{jia2025ecmrn} (54.6\%)}}
  {orange}{0.38}{0.18}  {1cm}{1.5cm}
&
\ZoomCellSpyCircle{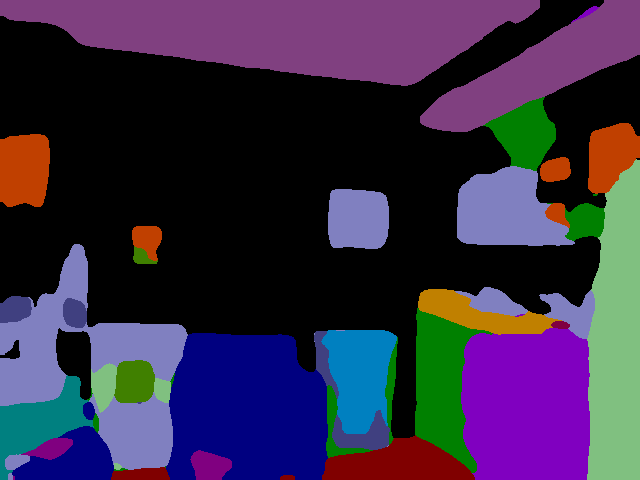}{0.162\textwidth}
  {\labA{MidLabel}{DFNet~\cite{gong2025diffpixelformer} (56.1\%)}}
  {orange}{0.38}{0.18}  {1cm}{1.5cm}
&
\ZoomCellSpyCircle{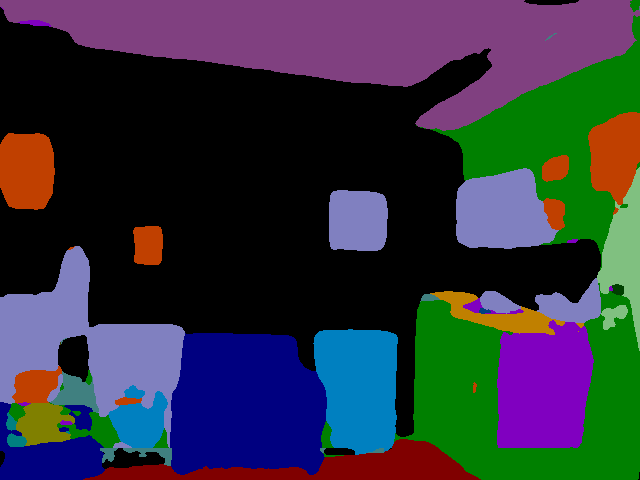}{0.162\textwidth}
  {\labA{OursLabel2}{\textbf{RSGMamba-S (56.4\%)}}}
  {orange}{0.38}{0.18}  {1cm}{1.5cm}
\\
\ZoomCellSpyCircle{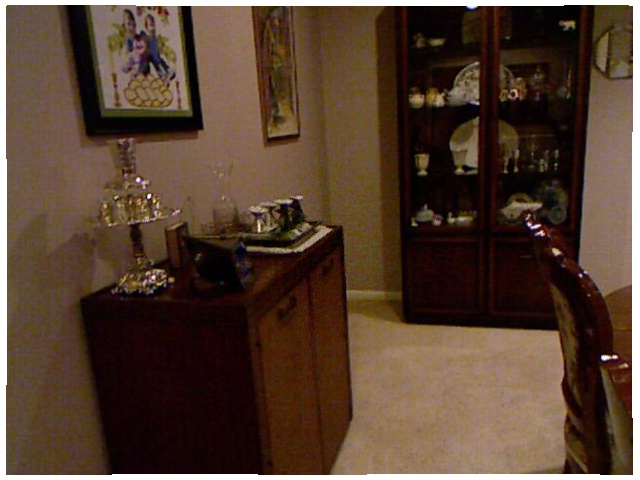}{0.162\textwidth}
  {\labA{MidLabel} {RGB img}}
  {cyan!60!black}{0.2}{0.55}  {1cm}{1.5cm}
&
\ZoomCellSpyCircle{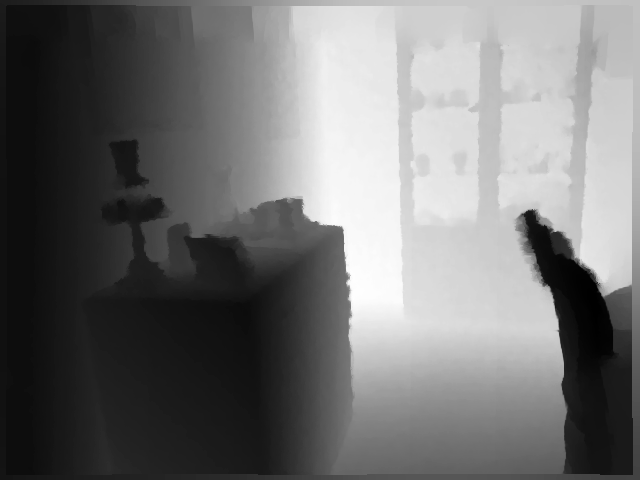}{0.162\textwidth}
  {\labA{MidLabel}{Depth img}}
  {orange}{0.2}{0.55}  {1cm}{1.5cm}
&
\ZoomCellSpyCircle{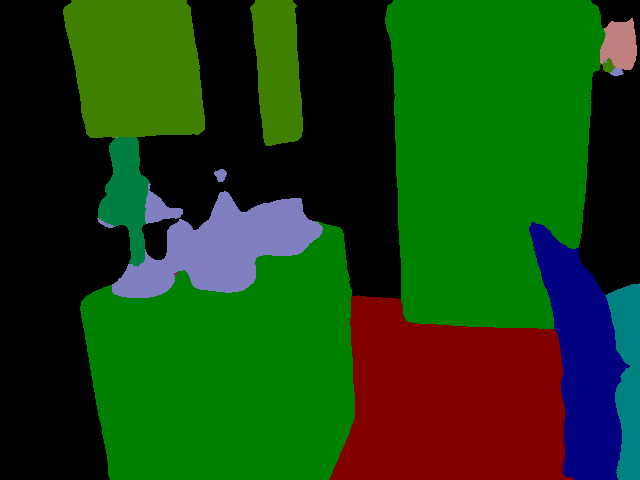}{0.162\textwidth}
  {\labA{MidLabel}{CMNext~\cite{zhang2023delivering} (56.9\%)}}
  {orange}{0.2}{0.55}  {1cm}{1.5cm}
&
\ZoomCellSpyCircle{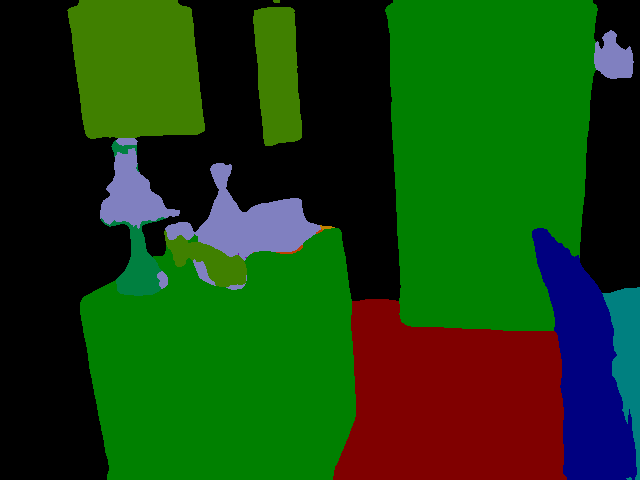}{0.162\textwidth}
  {\labA{MidLabel}{PrimKD~\cite{hao2024primkd} (57.8\%)}}
  {orange}{0.2}{0.55}  {1cm}{1.5cm}
&
\ZoomCellSpyCircle{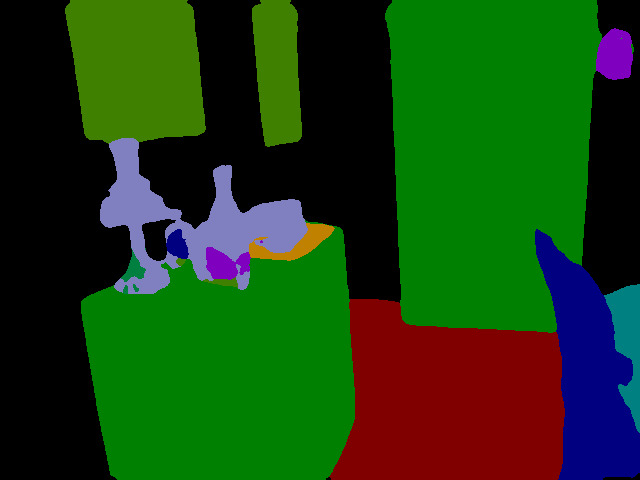}{0.162\textwidth}
  {\labA{MidLabel}{DFormerV2~\cite{yin2025dformerv2} (58.4\%)}}
  {orange}{0.2}{0.55}  {1cm}{1.5cm}
&
\ZoomCellSpyCircle{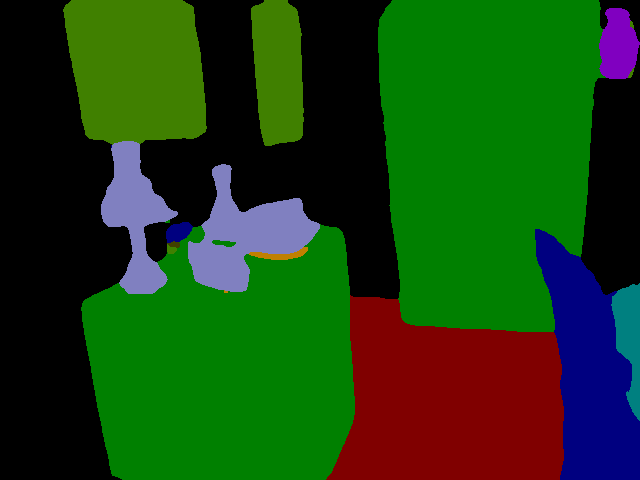}{0.162\textwidth}
  {\labA{OursLabel2}{\textbf{RSGMamba-B (58.8\%)}}}
  {orange}{0.2}{0.55}  {1cm}{1.5cm}

\end{tabular}

\begin{tabular}{@{}cccccc@{}}
\ZoomCellSpyCircle{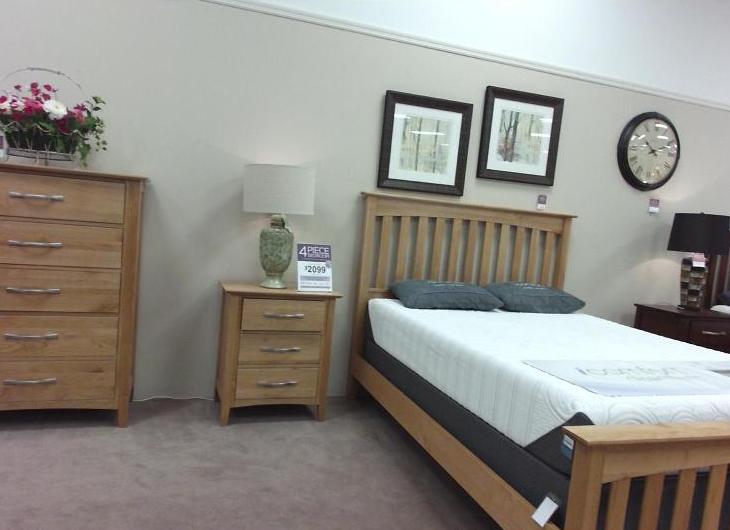}{0.162\textwidth}
  {\labA{MidLabel} {RGB img}}
  {cyan!60!black}{0.38}{0.32}  {1cm}{1.5cm}
&
\ZoomCellSpyCircle{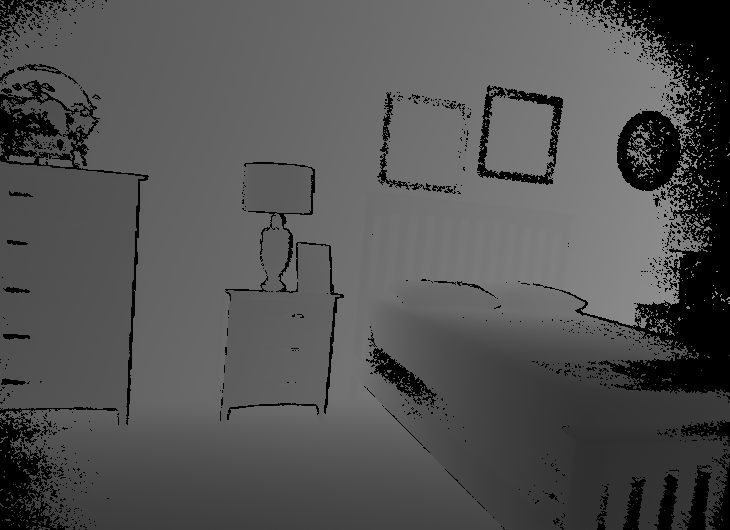}{0.162\textwidth}
  {\labA{MidLabel}{Depth img}}
  {orange}{0.38}{0.32}  {1cm}{1.5cm}
&
\ZoomCellSpyCircle{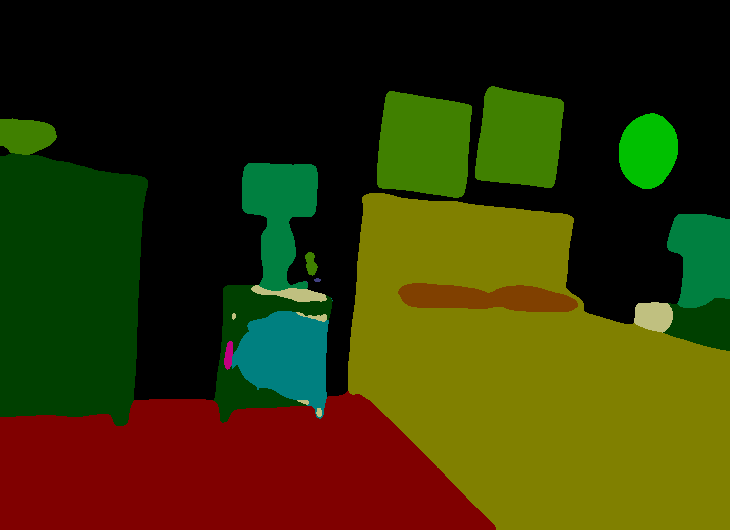}{0.162\textwidth}
  {\labA{MidLabel}{ECMRN~\cite{jia2025ecmrn} (50.5\%)}}
  {orange}{0.38}{0.32}  {1cm}{1.5cm}
&
\ZoomCellSpyCircle{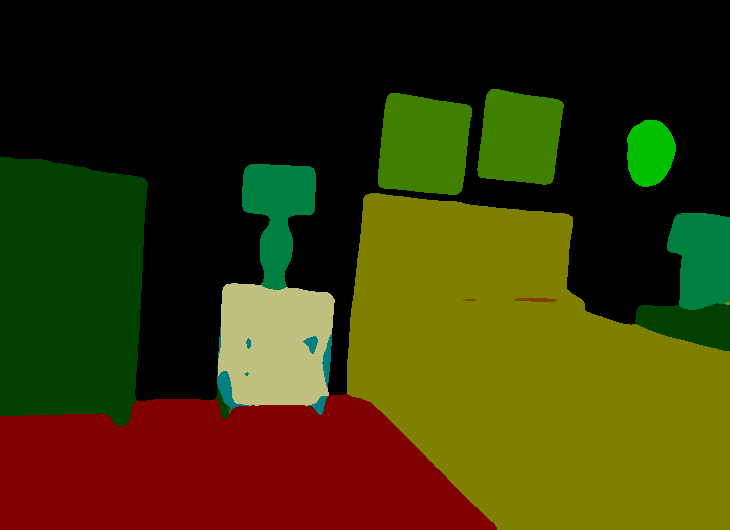}{0.162\textwidth}
  {\labA{MidLabel}{TokenFusion~\cite{kim2024token} (51.0\%)}}
  {orange}{0.38}{0.32}  {1cm}{1.5cm}
&
\ZoomCellSpyCircle{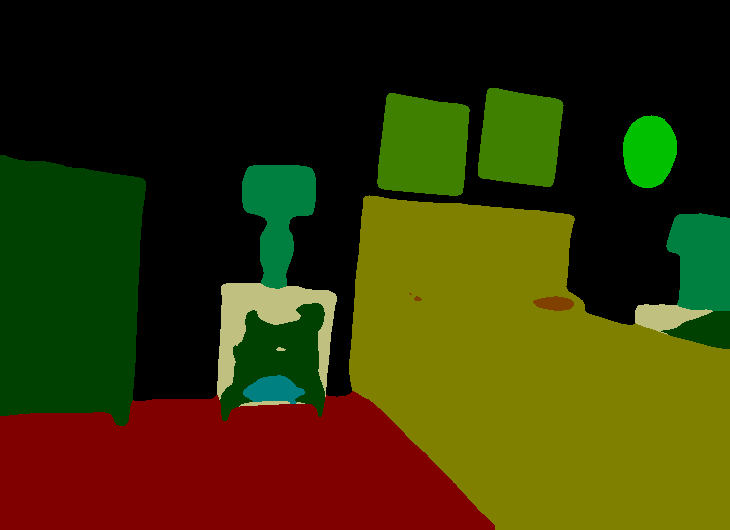}{0.162\textwidth}
  {\labA{MidLabel}{CMX~\cite{zhang2023cmx} (52.1\%)}}
  {orange}{0.38}{0.32}  {1cm}{1.5cm}
&
\ZoomCellSpyCircle{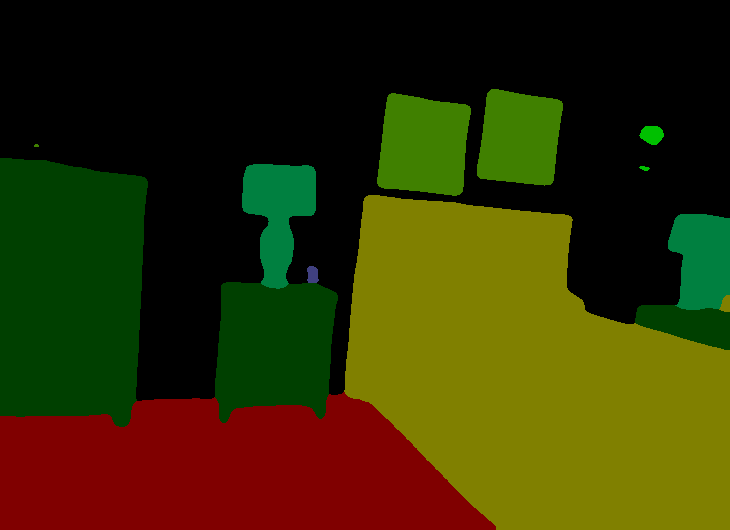}{0.162\textwidth}
  {\labA{OursLabel2}{\textbf{RSGMamba-S (52.9\%)}}}
  {orange}{0.38}{0.32}  {1cm}{1.5cm}
\\
\ZoomCellSpyCircle{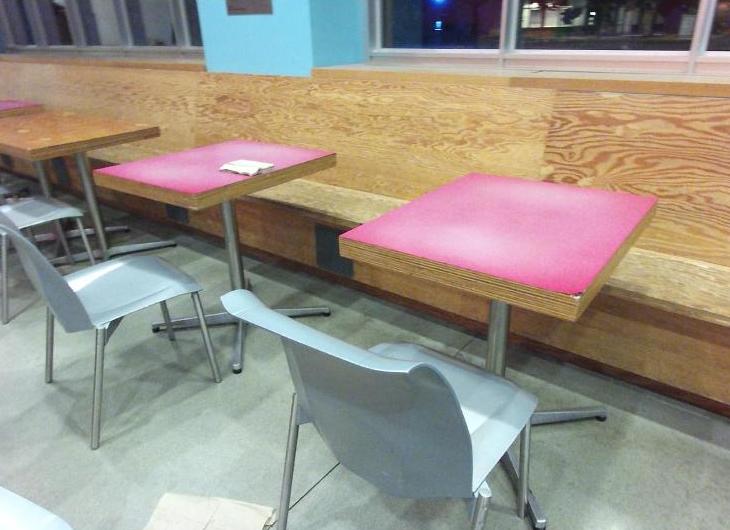}{0.162\textwidth}
  {\labA{MidLabel} {RGB img}}
  {cyan!60!black}{0.55}{0.55}  {1cm}{1.5cm}
&
\ZoomCellSpyCircle{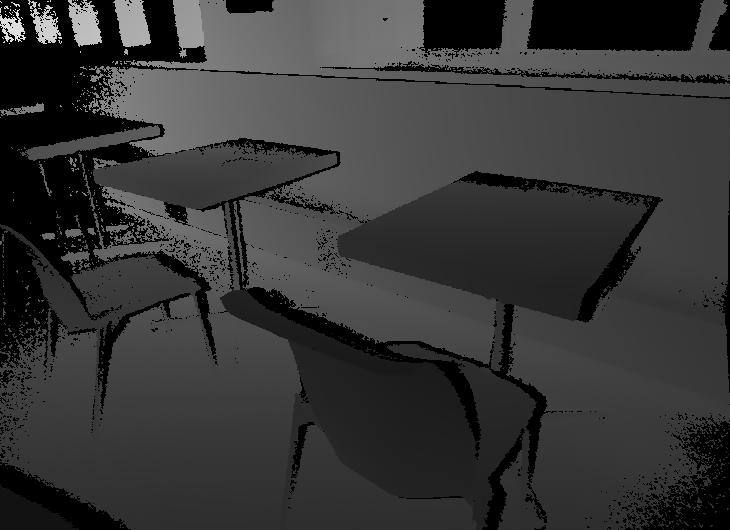}{0.162\textwidth}
  {\labA{MidLabel}{Depth img}}
  {orange}{0.55}{0.55}  {1cm}{1.5cm}
&
\ZoomCellSpyCircle{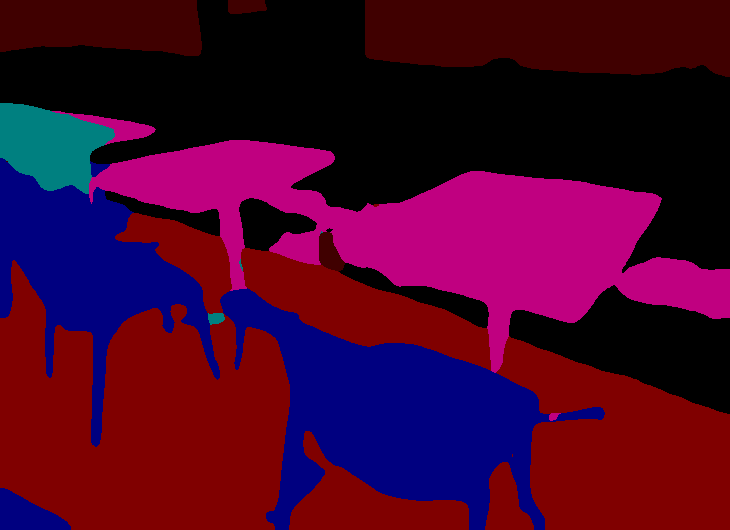}{0.162\textwidth}
  {\labA{MidLabel}{DFNet~\cite{yang2025difference} (52.4\%)}}
  {orange}{0.55}{0.55}  {1cm}{1.5cm}
&
\ZoomCellSpyCircle{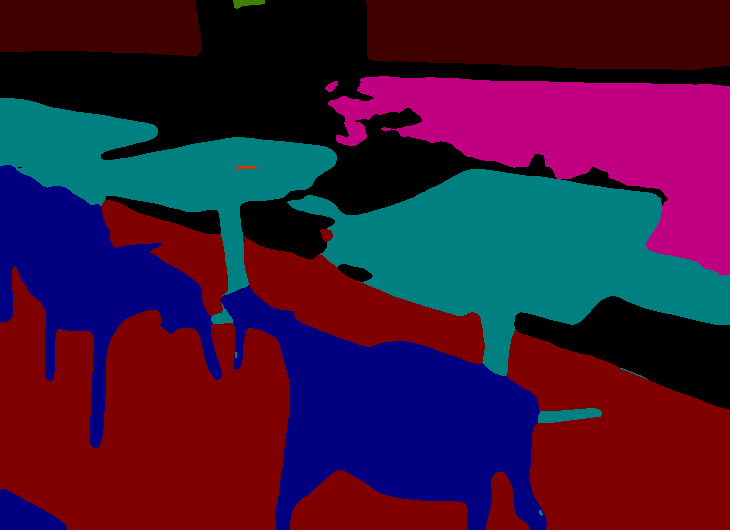}{0.162\textwidth}
  {\labA{MidLabel}{GeminiFusion~\cite{jia2024geminifusion} (52.7\%)}}
  {orange}{0.55}{0.55}  {1cm}{1.5cm}
&
\ZoomCellSpyCircle{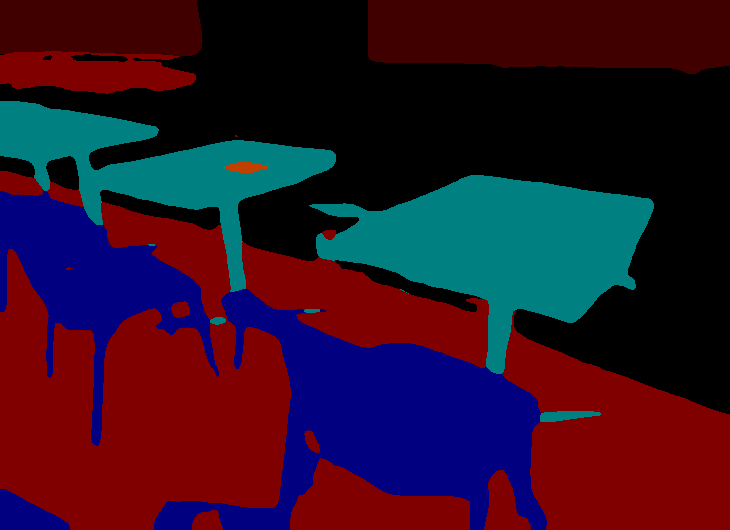}{0.162\textwidth}
  {\labA{MidLabel}{DFormerV2~\cite{yin2025dformerv2} (53.3\%)}}
  {orange}{0.55}{0.55}  {1cm}{1.5cm}
&
\ZoomCellSpyCircle{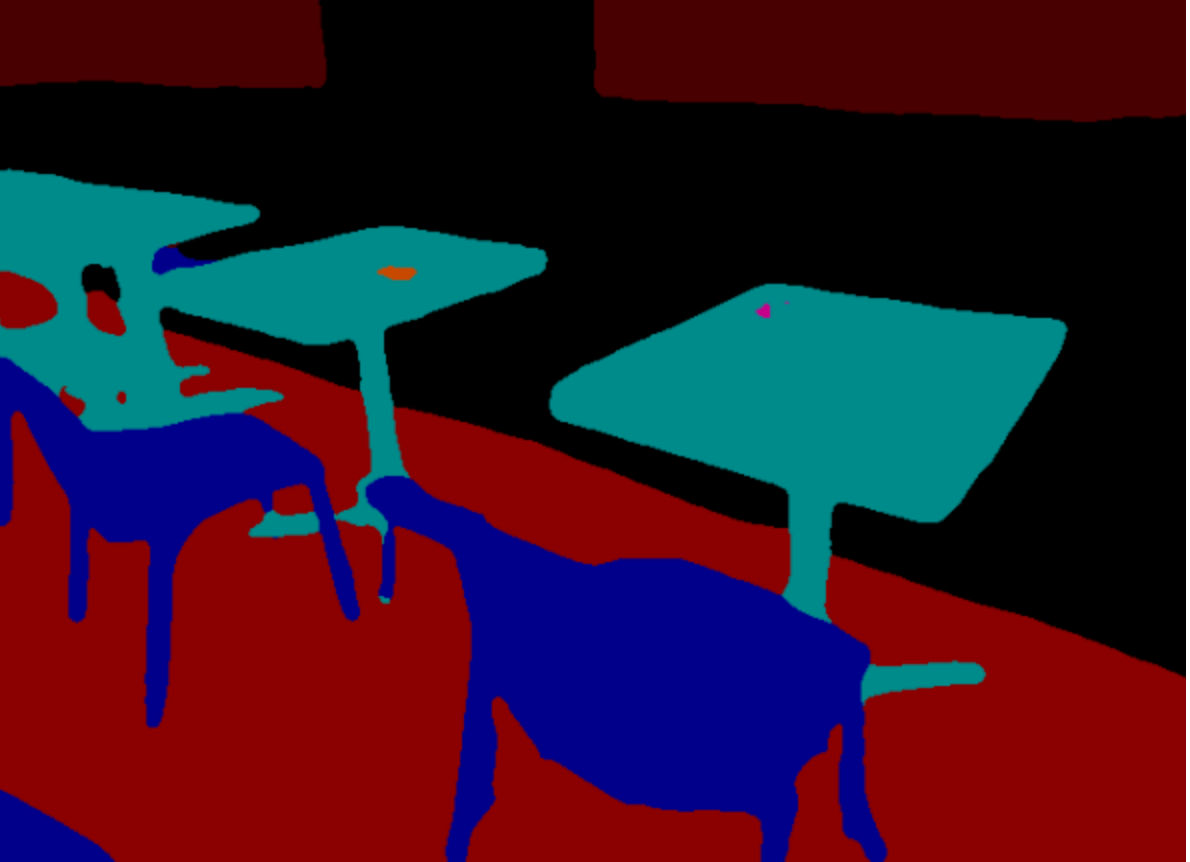}{0.162\textwidth}
  {\labA{OursLabel2}{\textbf{RSGMamba-B (54.0\%)}}}
  {orange}{0.55}{0.55}  {1cm}{1.5cm}
\end{tabular}

\caption{Qualitative comparisons on the NYUDepth V2~\cite{icra_2019_fastdepth} (first two rows) and SUN-RGBD~\cite{song2015sun} (last two rows) datasets. The highlighted regions (orange circles) emphasize the key differences among the methods. Compared with previous approaches, our RSGMamba produces more accurate predictions with clearer boundaries and better preservation of fine structures, especially in challenging regions with ambiguous depth or small objects. The mIoU (\%) for each method is reported below the corresponding result.}
\label{res:nyu & sun}
\end{figure*}

\section{Experiments}
\label {sect:experiment}

\subsection{Experimental Settings}

To comprehensively assess the effectiveness of RSGMamba, we conduct experiments on several widely used multimodal semantic segmentation benchmarks. Specifically, our method is tested on two RGB–Depth datasets, NYUDepth V2~\cite{icra_2019_fastdepth} and SUN-RGBD~\cite{song2015sun}, as well as two RGB–Thermal datasets, MFNet~\cite{ha2017mfnet} and PST900~\cite{shivakumar2020pst900}. These benchmarks cover diverse indoor and outdoor scenes with different modality characteristics, providing a rigorous testbed for cross-modal perception. Details of the datasets are provided below:
\begin{itemize}
    \item NYUDepth V2~\cite{icra_2019_fastdepth} consists of 1,449 paired RGB–D images captured at a resolution of $ 640 \times 480$. The dataset is split into 795 samples for training and 654 samples for evaluation, covering a total of 40 semantic classes.
    \item SUN-RGBD~\cite{song2015sun} contains 10,335 RGB–D image pairs annotated with 37 object categories. Following the standard split, 5,285 images are used for training, and the remaining 5,050 images are used for testing.
    \item MFNet~\cite{ha2017mfnet} dataset consists of paired RGB–Thermal images captured under both daytime (820 images) and nighttime (749 images) conditions, all with a spatial resolution of $ 640 \times 480$. It covers eight object categories commonly encountered in road and driving scenes.
    \item PST900~\cite{shivakumar2020pst900} dataset contains 894 paired RGB–Thermal images collected in complex underground scenes. It offers pixel-level semantic annotations for four object categories. Both the RGB and thermal modalities are available at a resolution of $1280 \times 720$. The dataset is split into training and testing subsets using a 2:1 ratio.
\end{itemize}

\subsection{Training Settings}
To ensure a fair and consistent comparison with existing approaches, we adopt the same data pre-processing pipeline and augmentation protocols as those employed in previous works.
For optimization, we adopt the AdamW optimizer~\cite{loshchilov2017decoupled} with a weight decay of 0.01. The initial learning rate is set to $6 \times 10^{-5}$, and the model is trained for 500 epochs with a batch size of 8. A polynomial learning rate decay schedule is employed, along with a warm-up phase during the first 10 epochs to stabilize training. The network is optimized using the standard cross-entropy loss. All experiments are conducted on four NVIDIA H100 GPUs.

\subsection{State-of-the-art Comparisons}
\textbf{RGB-D Segmentation} Table~\ref{RGBD} presents comprehensive comparisons between our method and existing state-of-the-art RGB-D semantic segmentation approaches on the NYUDepth V2 and SUN-RGBD benchmarks. RSGMamba-S (SegMAN-S) achieves 56.4\% mIoU on NYUDepth V2 and 52.9\% on SUN-RGBD, performing on par with DiffPixelFormer~\cite{gong2025diffpixelformer} (56.3\% / 52.8\%). However, this comparable performance is obtained with only 28.10M parameters and 58.45G FLOPs, whereas DiffPixelFormer~\cite{gong2025diffpixelformer} requires 85.4M parameters and 154.8G FLOPs under the same input resolution. In other words, RSGMamba-S reduces the parameter count by more than 2.6$\times$ and the computational cost by nearly 3$\times$, while maintaining equivalent segmentation accuracy. Compared with DFormerV2-L~\cite{yin2025dformerv2}, our RSGMamba-B (SegMAN-B) consistently achieves superior performance under comparable computational budgets. As reported in Table~\ref{RGBD}, both methods operate at similar FLOPs levels on NYUDepth V2 (130.2G vs. 124.1G) and SUN-RGBD (168.4G vs. 160.5G). Despite this, RSGMamba-B employs only 48.55M parameters, which is nearly half that of DFormerV2-L~\cite{yin2025dformerv2} (95.5M parameters), while delivering higher segmentation accuracy on both benchmarks (58.8\% vs. 58.4\% on NYUDepth V2 and 54.0\% vs. 53.3\% on SUN-RGBD). As illustrated in Fig.~\ref{res:nyu & sun}, our method produces more accurate predictions with clearer boundaries and better preservation of fine structures.

\begin{figure*}[!t]
\centering
\setlength{\tabcolsep}{0pt}
\renewcommand{\arraystretch}{1.0}

\begin{tabular}{@{}cccccc@{}}
\ZoomCellSpyCircle{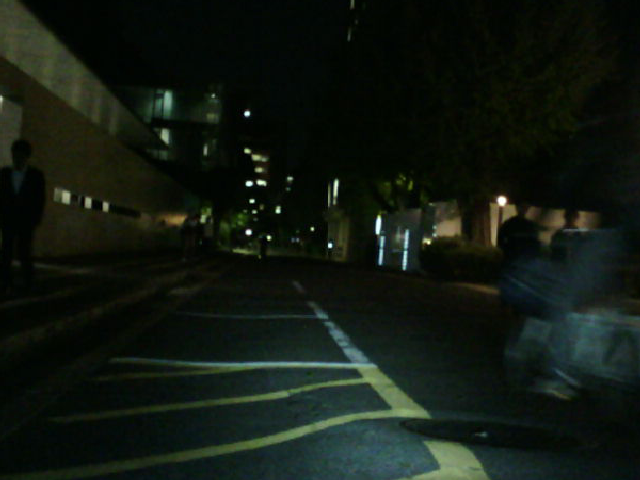}{0.162\textwidth}
  {\labA{MidLabel} {RGB img}}
  {cyan!60!black}{0.9}{0.4}  {1cm}{1.5cm}
&
\ZoomCellSpyCircle{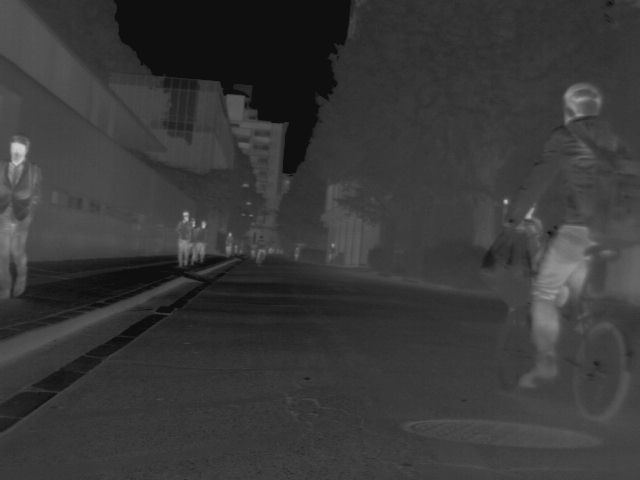}{0.162\textwidth}
  {\labA{MidLabel}{Thermal img}}
  {orange}{0.9}{0.4}  {1cm}{1.5cm}
&
\ZoomCellSpyCircle{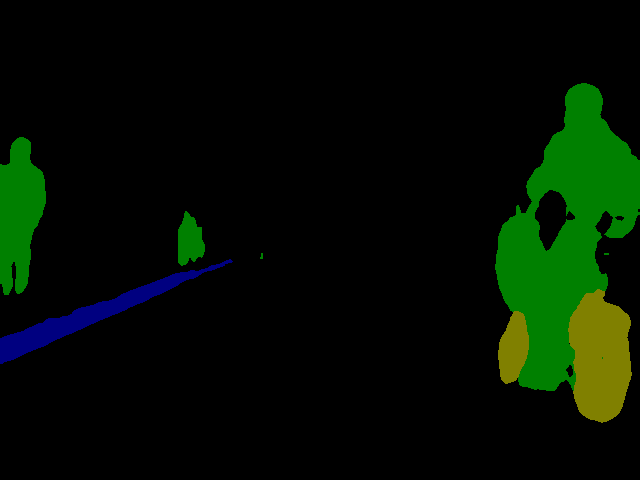}{0.162\textwidth}
  {\labA{MidLabel}{HKDNet~\cite{tu2025hybrid} (56.5\%)}}
  {orange}{0.9}{0.4}  {1cm}{1.5cm}
&
\ZoomCellSpyCircle{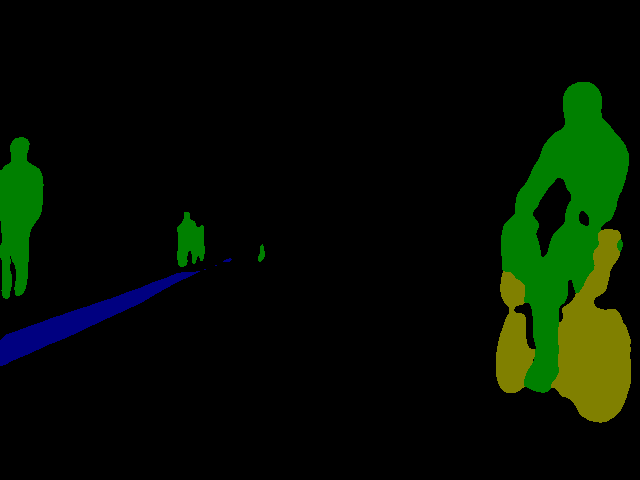}{0.162\textwidth}
  {\labA{MidLabel}{AGFNet~\cite{zhou2025agfnet} (58.6\%)}}
  {orange}{0.9}{0.4}  {1cm}{1.5cm}
&
\ZoomCellSpyCircle{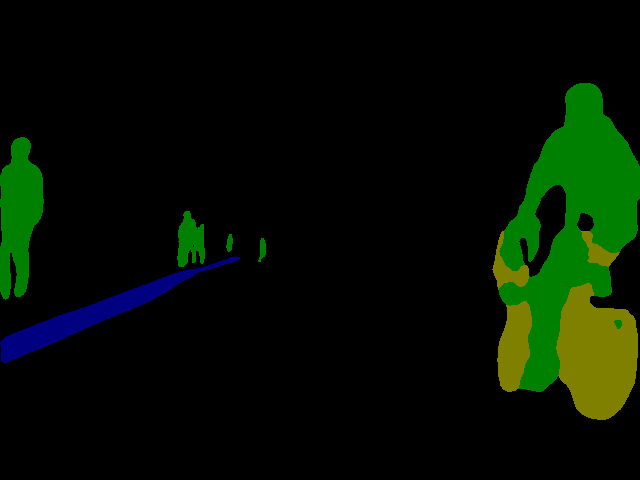}{0.162\textwidth}
  {\labA{MidLabel}{CRM~\cite{shin2024complementary} (59.1\%)}}
  {orange}{0.9}{0.4}  {1cm}{1.5cm}
&
\ZoomCellSpyCircle{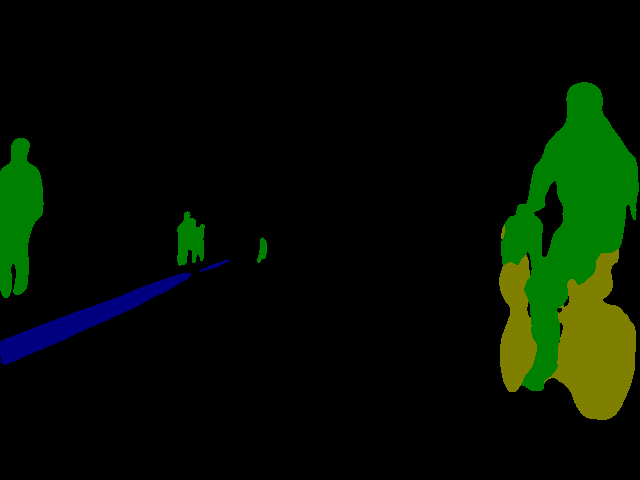}{0.162\textwidth}
  {\labA{OursLabel2}{\textbf{RSGMamba-S (59.4\%)}}}
  {orange}{0.9}{0.4}  {1cm}{1.5cm}
\\
\ZoomCellSpyCircle{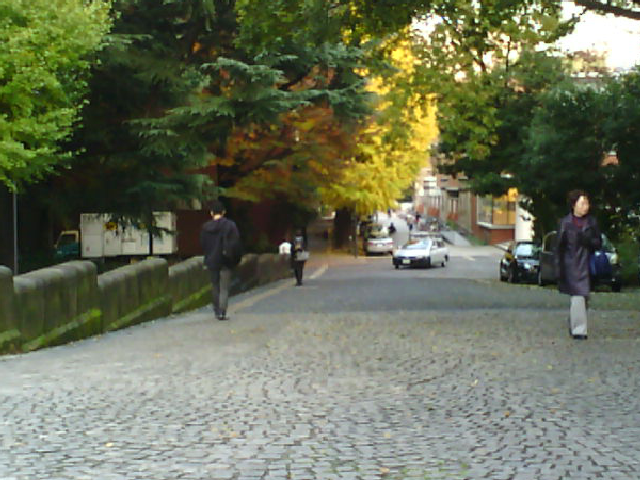}{0.162\textwidth}
  {\labA{MidLabel} {RGB img}}
  {cyan!60!black}{0.65}{0.5}  {1cm}{1.5cm}
&
\ZoomCellSpyCircle{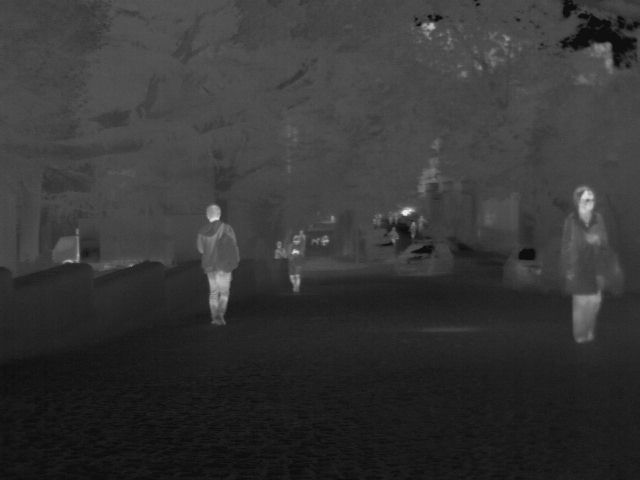}{0.162\textwidth}
  {\labA{MidLabel}{Thermal img}}
  {orange}{0.65}{0.5}  {1cm}{1.5cm}
&
\ZoomCellSpyCircle{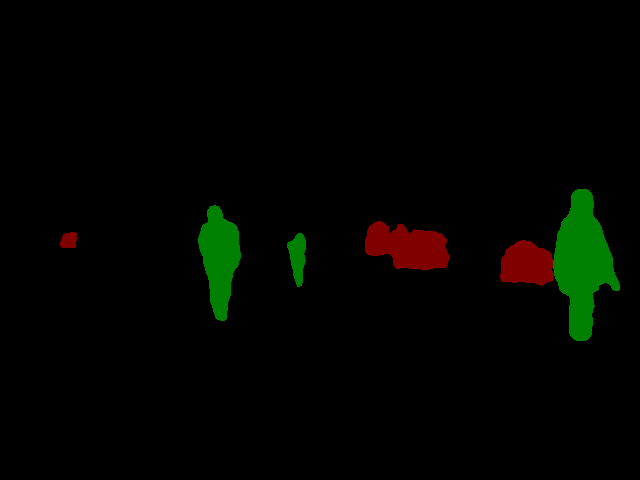}{0.162\textwidth}
  {\labA{MidLabel}{OmniFuse~\cite{zhang2025omnifuse} (56.3\%)}}
  {orange}{0.65}{0.5}  {1cm}{1.5cm}
&
\ZoomCellSpyCircle{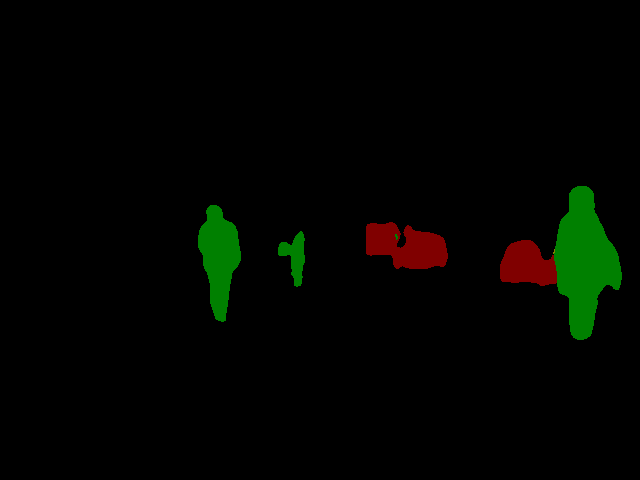}{0.162\textwidth}
  {\labA{MidLabel}{CMNeXt~\cite{zhang2023delivering} (59.9\%)}}
  {orange}{0.65}{0.5}  {1cm}{1.5cm}
&
\ZoomCellSpyCircle{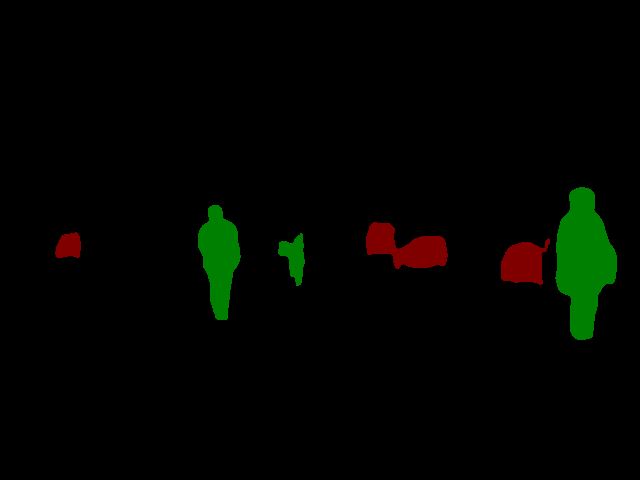}{0.162\textwidth}
  {\labA{MidLabel}{Mixprompt~\cite{haomixprompt} (60.1\%)}}
  {orange}{0.65}{0.5}  {1cm}{1.5cm}
&
\ZoomCellSpyCircle{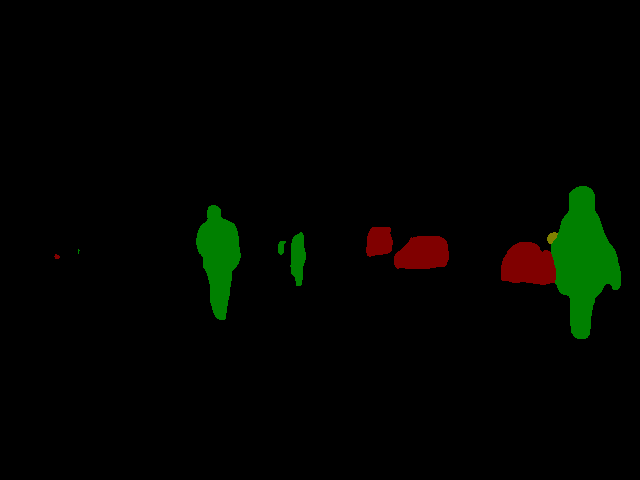}{0.162\textwidth}
  {\labA{OursLabel2}{\textbf{RSGMamba-B (61.1\%)}}}
  {orange}{0.65}{0.5}  {1cm}{1.5cm}
\end{tabular}

\begin{tabular}{@{}cccccc@{}}
\ZoomCellSpyCircle{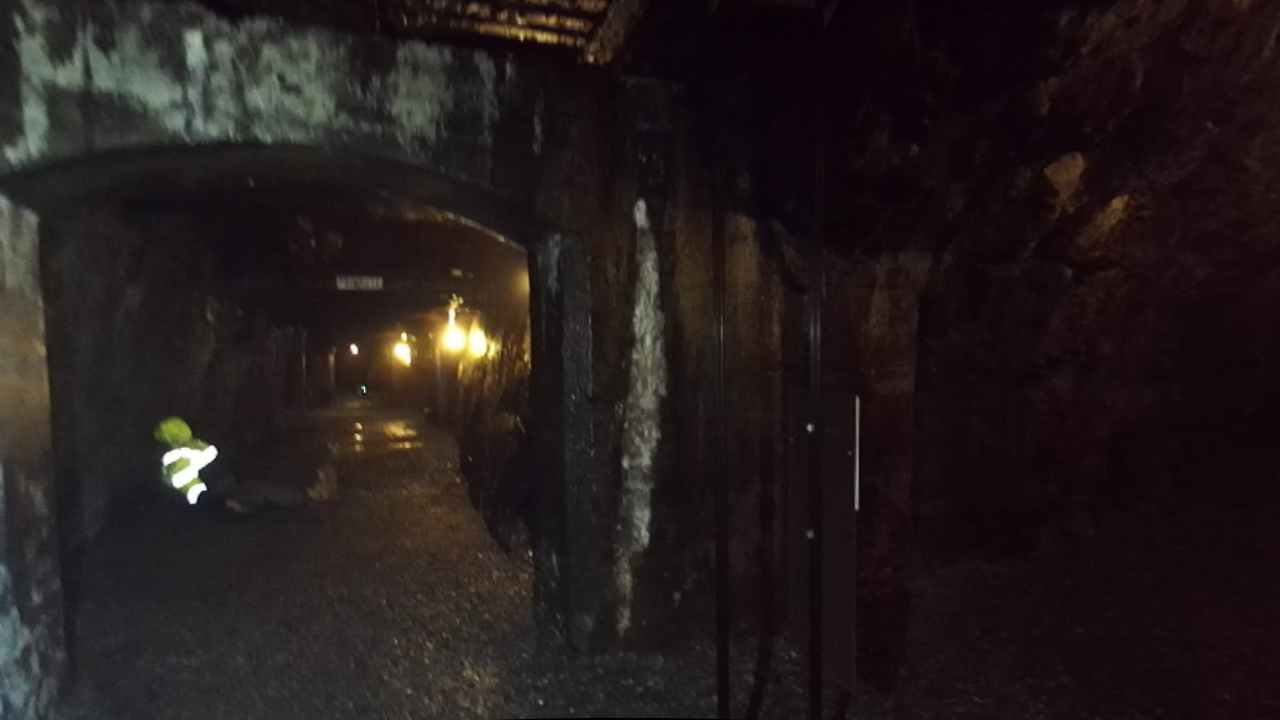}{0.162\textwidth}
  {\labA{MidLabel} {RGB img}}
  {cyan!60!black}{0.2}{0.4}  {1cm}{1.5cm}
&
\ZoomCellSpyCircle{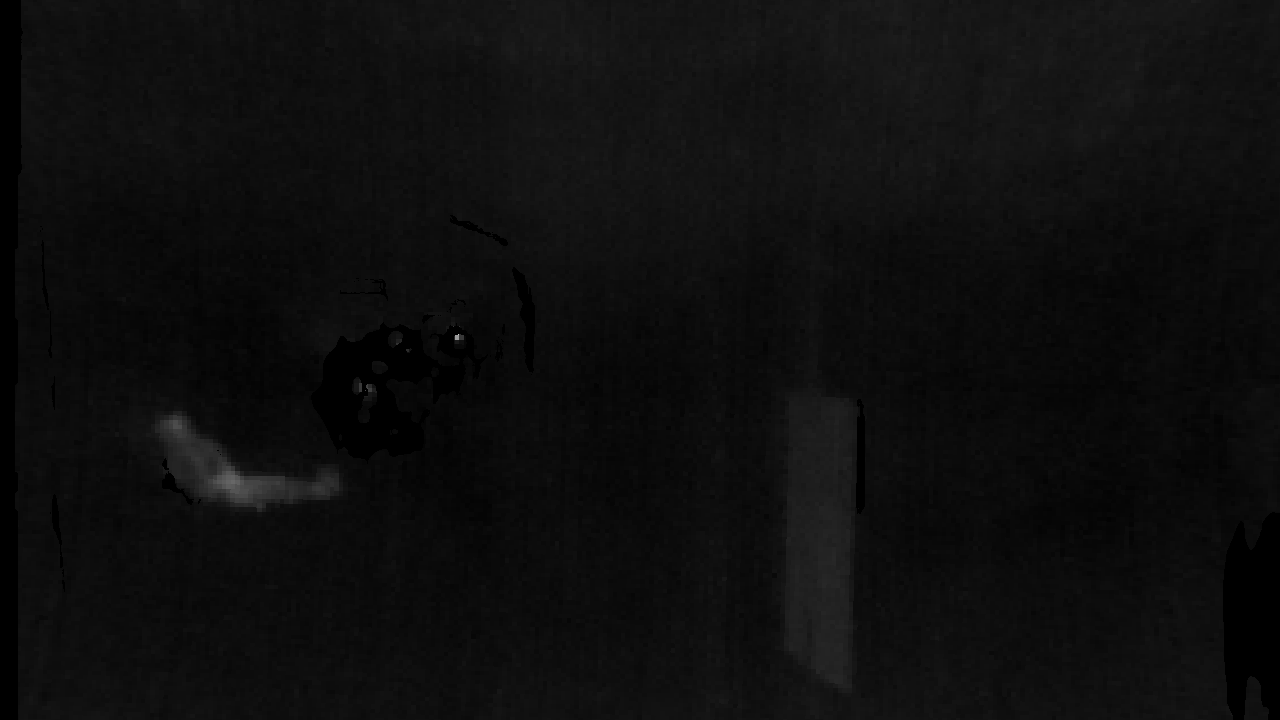}{0.162\textwidth}
  {\labA{MidLabel}{Thermal img}}
  {orange}{0.2}{0.4}  {1cm}{1.5cm}
&
\ZoomCellSpyCircle{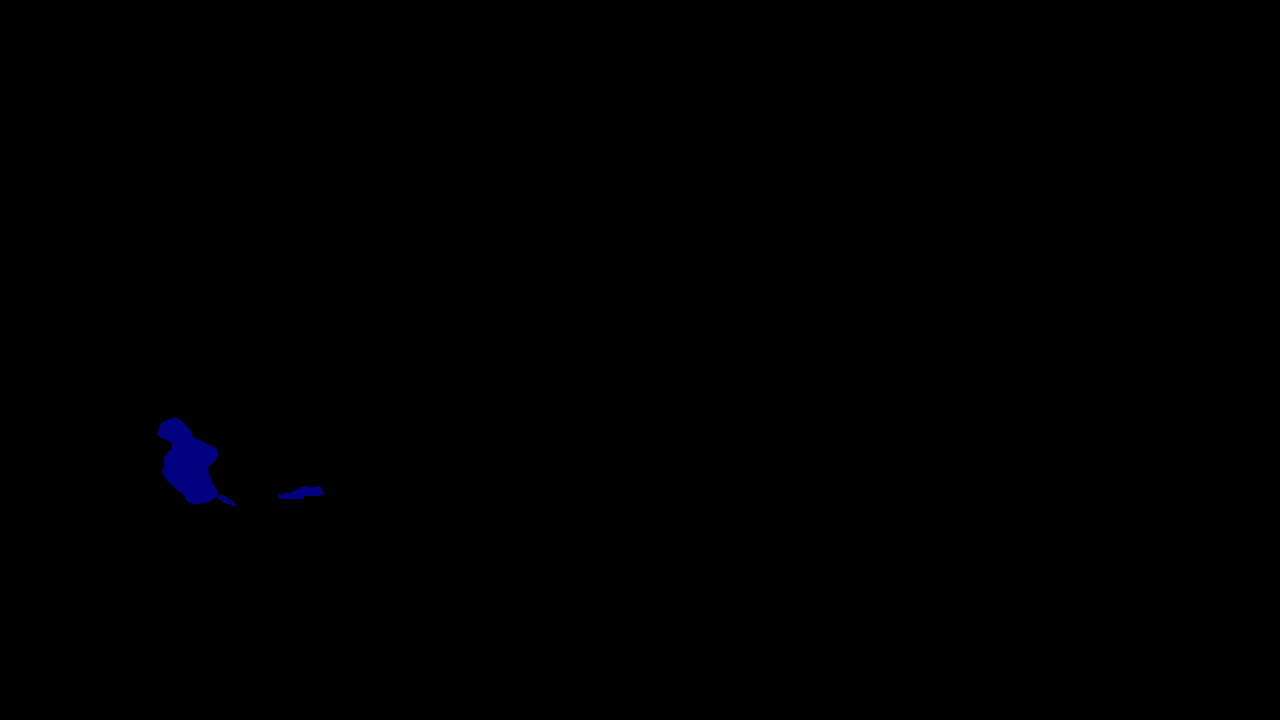}{0.162\textwidth}
  {\labA{MidLabel}{SGFNet~\cite{wang2023sgfnet} (82.8\%)}}
  {orange}{0.2}{0.4}  {1cm}{1.5cm}
&
\ZoomCellSpyCircle{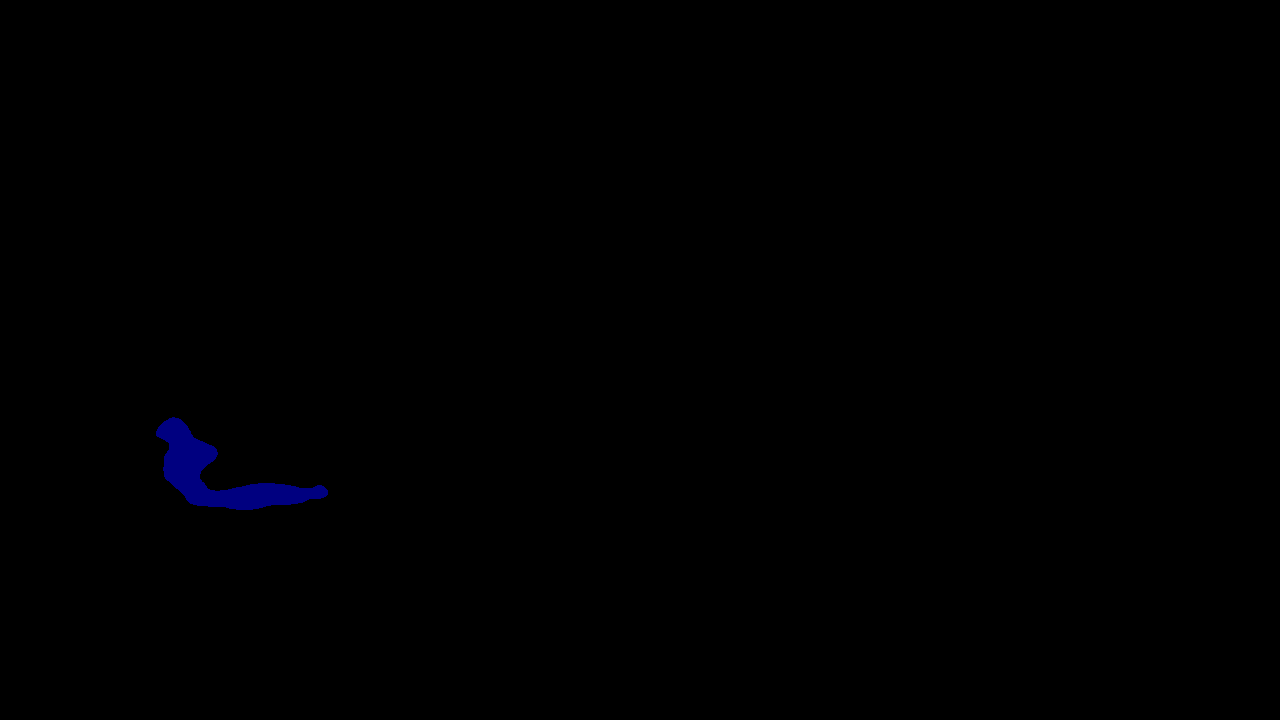}{0.162\textwidth}
  {\labA{MidLabel}{AGFNet~\cite{zhou2025agfnet} (84.8\%)}}
  {orange}{0.2}{0.4}  {1cm}{1.5cm}
&
\ZoomCellSpyCircle{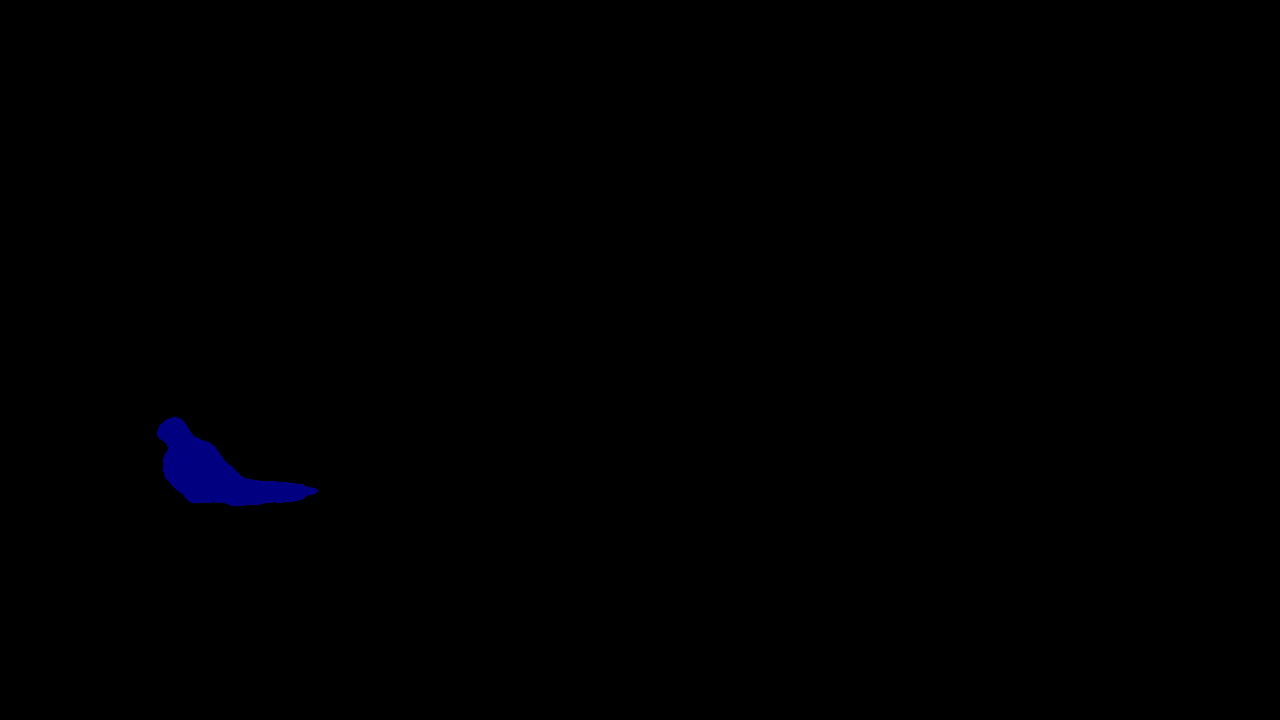}{0.162\textwidth}
  {\labA{MidLabel}{CPAL~\cite{liu2025cpal} (86.8\%)}}
  {orange}{0.2}{0.4}  {1cm}{1.5cm}
&
\ZoomCellSpyCircle{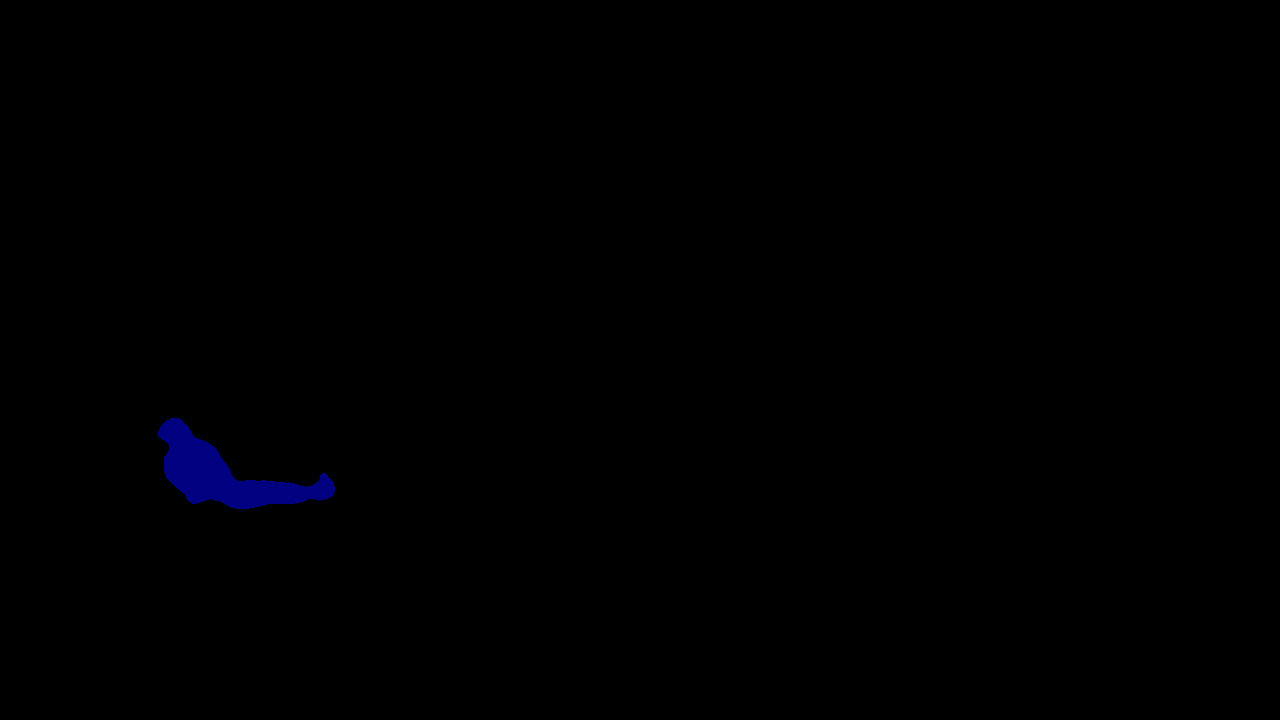}{0.162\textwidth}
  {\labA{OursLabel2}{\textbf{RSGMamba-S (87.3\%)}}}
  {orange}{0.2}{0.4}  {1cm}{1.5cm}
\\
\ZoomCellSpyCircle{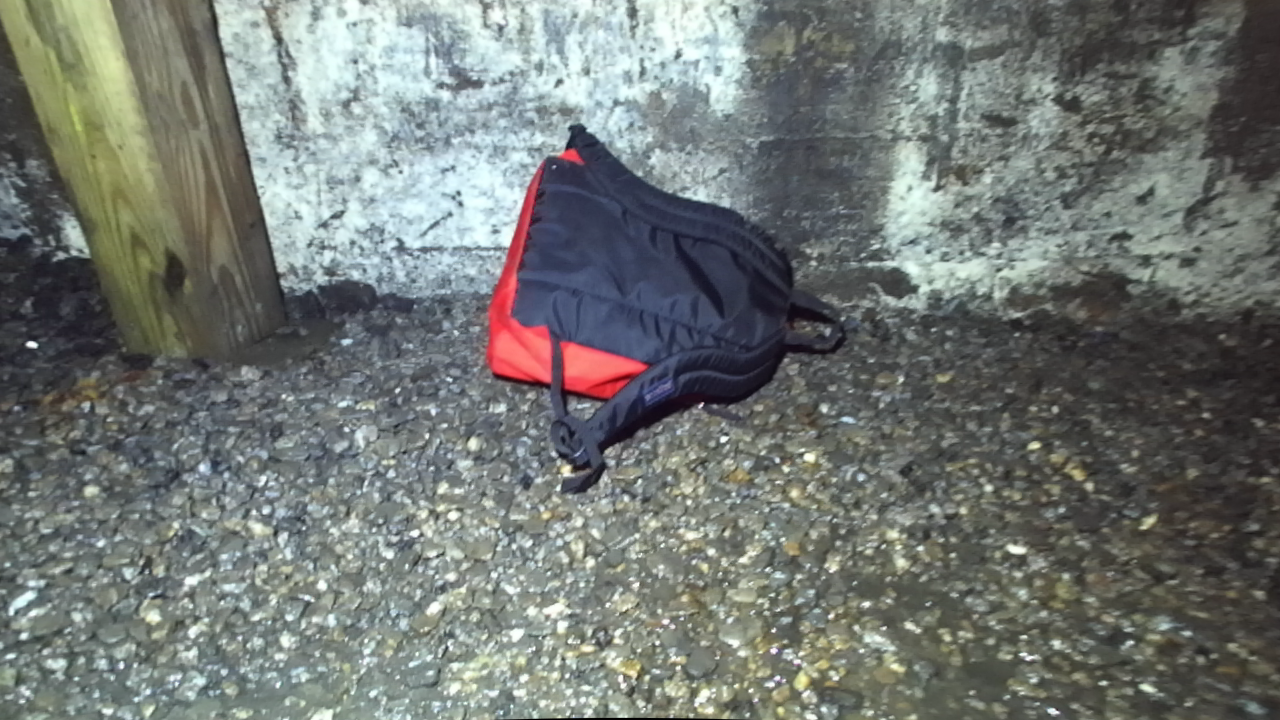}{0.162\textwidth}
  {\labA{MidLabel} {RGB img}}
  {cyan!60!black}{0.4}{0.5}  {1cm}{1.5cm}
&
\ZoomCellSpyCircle{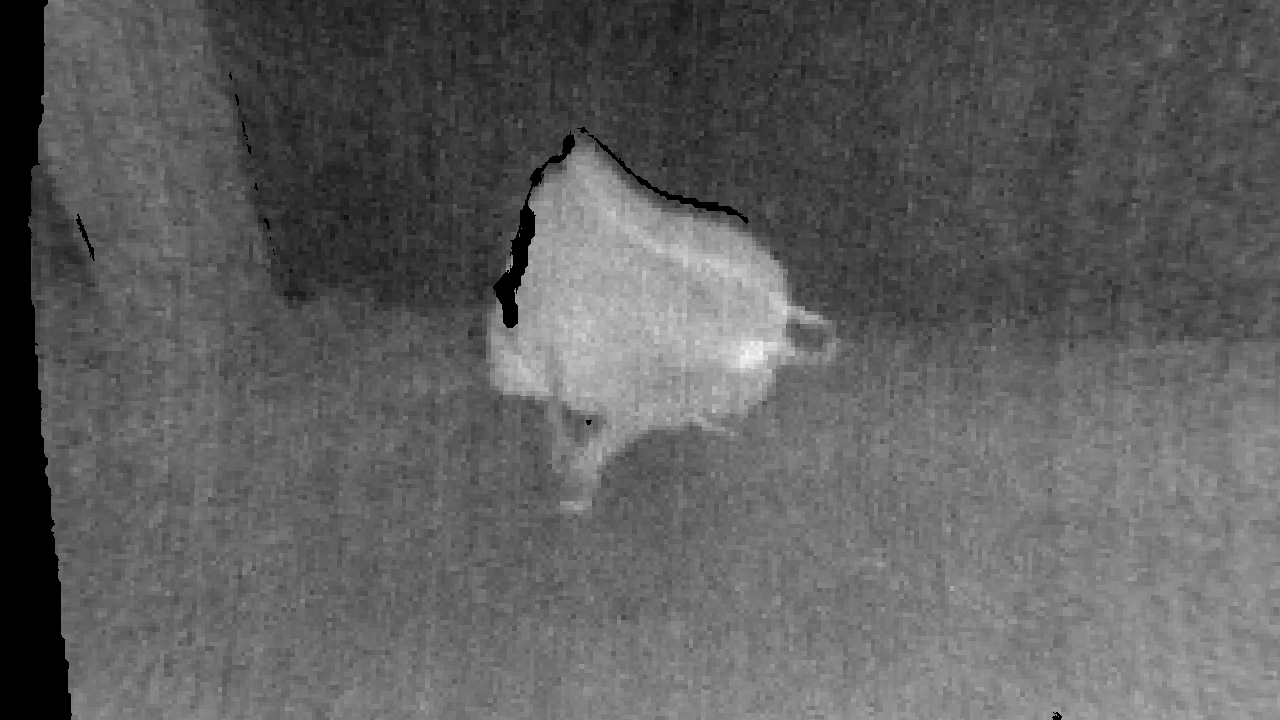}{0.162\textwidth}
  {\labA{MidLabel}{Thermal img}}
  {orange}{0.4}{0.5}  {1cm}{1.5cm}
&
\ZoomCellSpyCircle{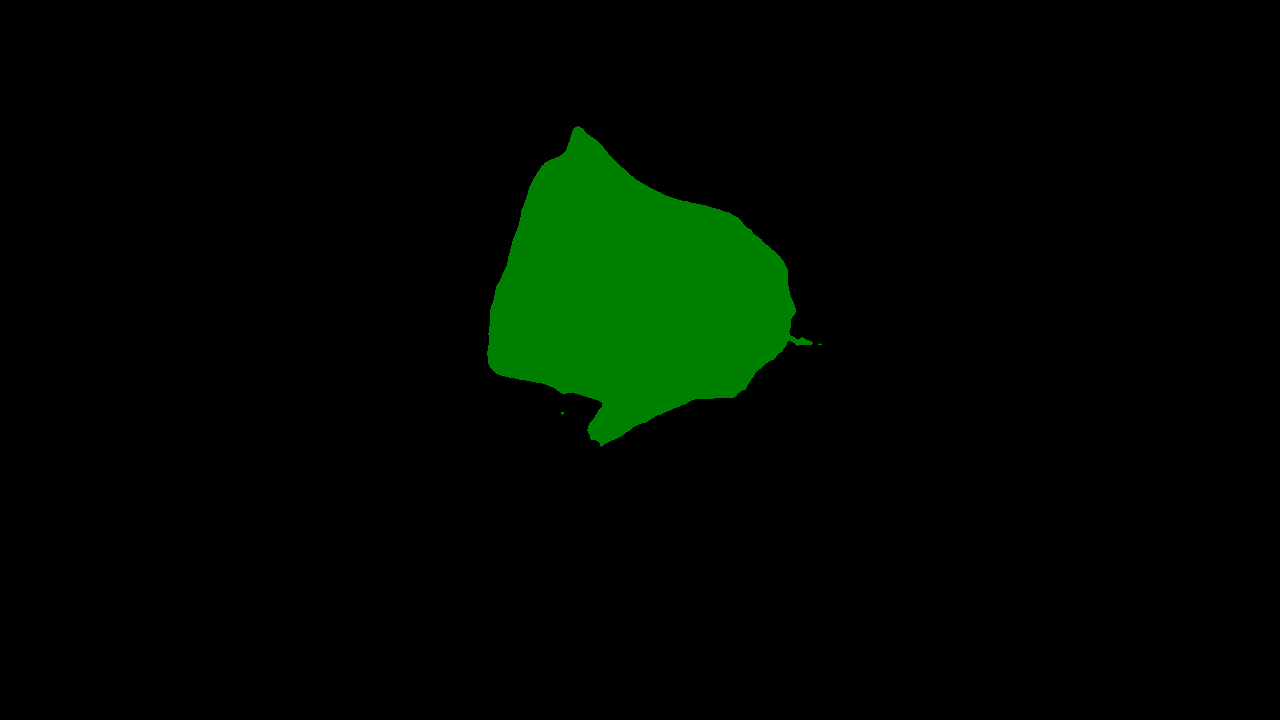}{0.162\textwidth}
  {\labA{MidLabel}{MiLNet~\cite{liu2025milnet}  (85.1\%)}}
  {orange}{0.4}{0.5}  {1cm}{1.5cm}
&
\ZoomCellSpyCircle{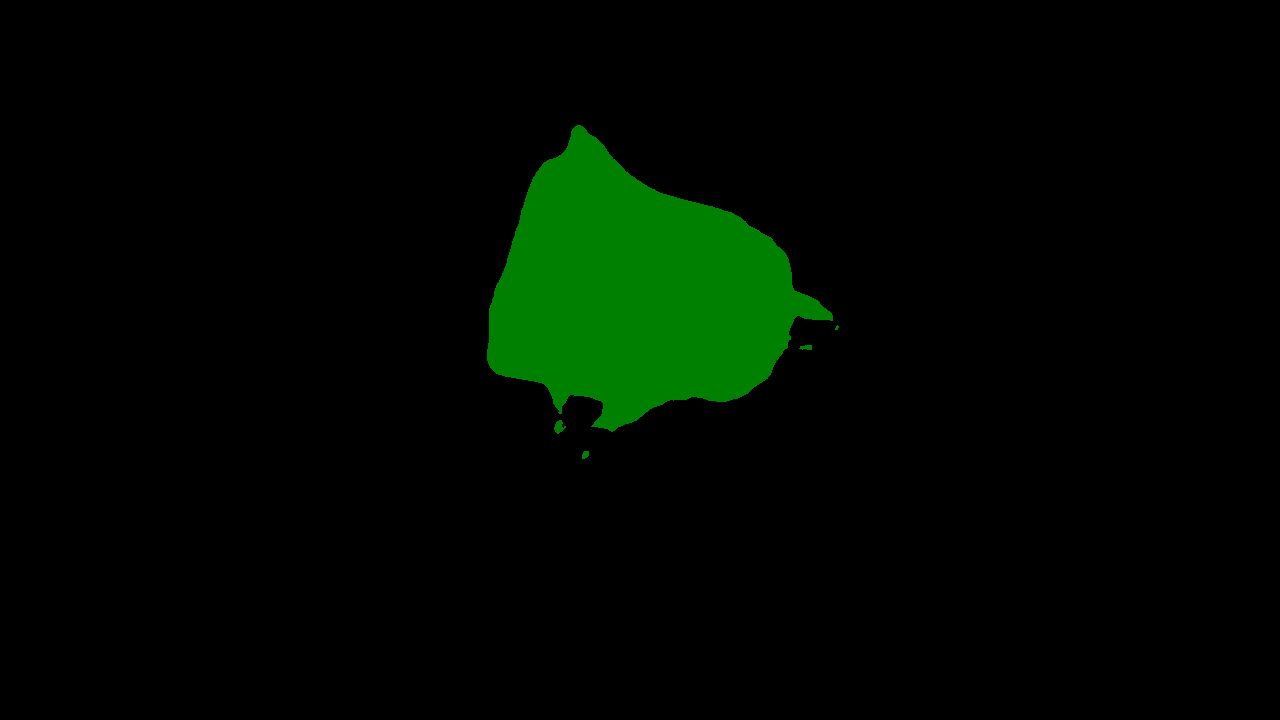}{0.162\textwidth}
  {\labA{MidLabel}{DPLNet~\cite{dong2024efficient}  (86.7\%)}}
  {orange}{0.4}{0.5}  {1cm}{1.5cm}
&
\ZoomCellSpyCircle{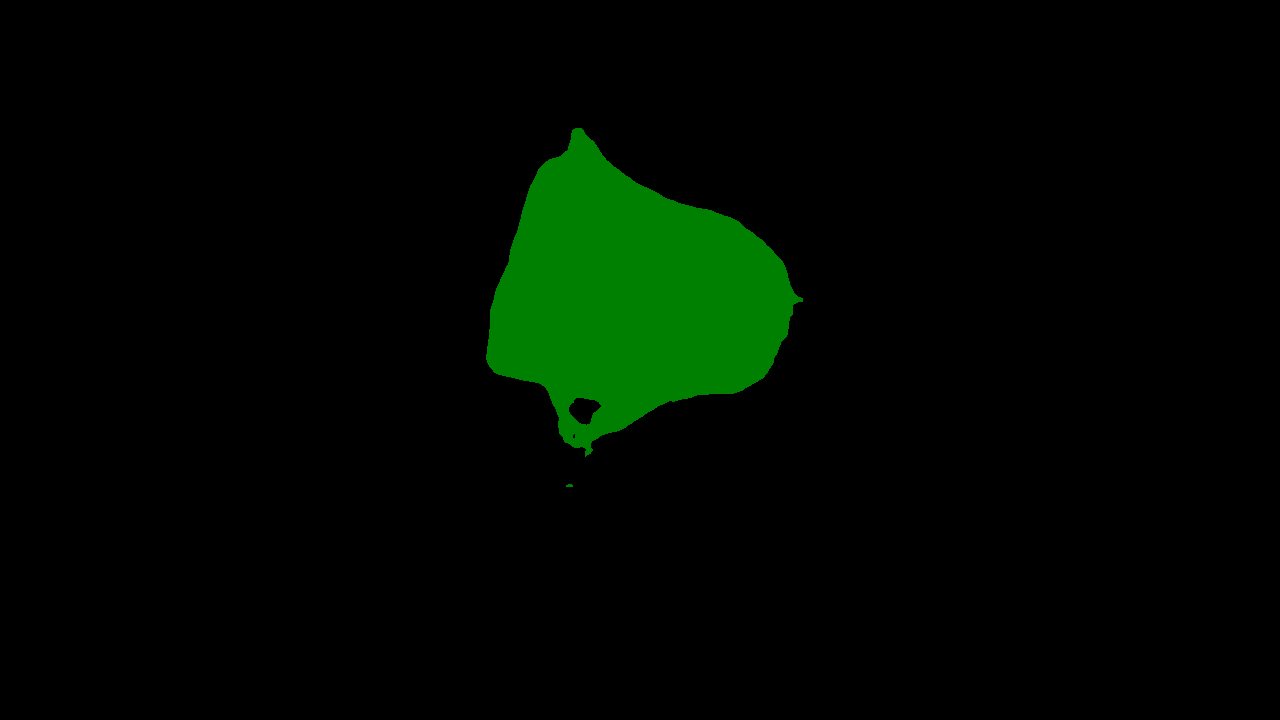}{0.162\textwidth}
  {\labA{MidLabel}{Sigma~\cite{wan2025sigma} (88.6\%)}}
  {orange}{0.4}{0.5}  {1cm}{1.5cm}
&
\ZoomCellSpyCircle{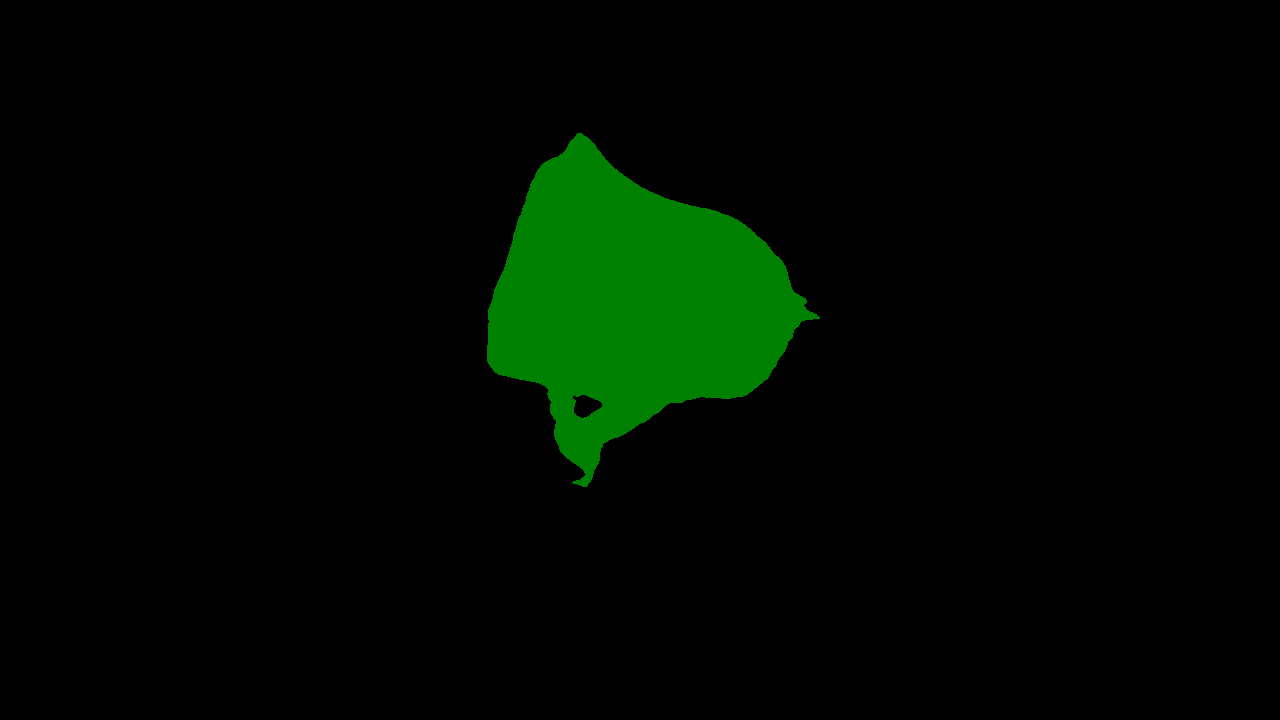}{0.162\textwidth}
  {\labA{OursLabel2}{\textbf{RSGMamba-B (88.9\%)}}}
  {orange}{0.4}{0.5}  {1cm}{1.5cm}
\end{tabular}

\caption{Qualitative comparisons on the MFNet~\cite{ha2017mfnet} (first two rows) and PST900~\cite{shivakumar2020pst900} (last two rows) datasets. The highlighted regions (orange circles) emphasize the key differences among methods under challenging conditions, such as low illumination and thermal ambiguity. Compared with previous approaches, our RSGMamba produces more accurate and consistent predictions, with clearer boundaries and better detection of small or hard-to-distinguish objects. The mIoU (\%) for each method is reported below the corresponding result.}
\label{res:mfnet & pst900}
\end{figure*}

\textbf{RGB-T Segmentation} Table~\ref{MFNet} presents quantitative comparisons on the MFNet~\cite{ha2017mfnet} RGB-T semantic segmentation benchmark. As shown, RSGMamba achieves the best overall performance across both model scales, demonstrating strong robustness under diverse illumination conditions. In particular, RSGMamba-B (SegMAN-B) attains the highest mIoU of 61.1\%, surpassing recent state-of-the-art methods such as Sigma~\cite{wan2025sigma} and CMX~\cite{zhang2023cmx} by a clear margin. These results validate the effectiveness of the proposed reliability-aware self-gated fusion mechanism in mitigating unreliable thermal cues and selectively enhancing complementary information, thereby leading to more accurate and stable segmentation in real-world RGB-T scenarios. In addition, as shown in Table~\ref{PST900}, our method outperforms competing methods by more than 0.3\% (88.9\% vs. 88.6\%) on the PST900 dataset, further demonstrating the superiority of our proposed method. As shown in Fig.~\ref{res:mfnet & pst900}, the highlighted regions further demonstrate that our method achieves more accurate and consistent predictions under challenging conditions such as low illumination and thermal ambiguity, with improved perception of small or hard-to-distinguish objects.

\begin{table*}[!t]
\begin{center}
\caption{Quantitative comparisons on the MFNet dataset. The \colorbox[RGB]{255,205,205}{best} and \colorbox[RGB]{255,231,198}{second} best results for each metric are color-coded.}

\label{MFNet}
\scalebox{1.0}{
\begin{tabular}{lcccccccccccc}
\toprule
\textbf{Method} & \textbf{Publication}  & \textbf{Backbone}  & \textbf{Unlabeled} & \textbf{Car} & \textbf{Person} & \textbf{Bike} & \textbf{Curve} & \textbf{Stop} & \textbf{Guardrail} & \textbf{Cone} & \textbf{Bump} & \textbf{mIoU} \\
\midrule 

ABMDRNet~\cite{zhang2021abmdrnet} &CVPR'21  & ResNet-50  & \cellcolor[RGB]{255,205,205}98.6 & 84.8 & 69.6 & 60.3 & 45.1 & 33.1 & 5.1 & 47.4 & 50.0 & 54.8 \\
GMNet~\cite{zhou2021gmnet} &TIP'21  & ResNet-50  & 97.5 & 86.5 & 73.1 & 61.7 & 44.0 & 42.3 & \cellcolor[RGB]{255,231,198}14.5 & 48.7 & 47.4 & 57.3 \\
LASNet~\cite{li2022rgb} & TCSVT'23 & RGB-T & 97.4 &84.2 & 67.1 & 56.9 & 41.1 & 39.6 & \cellcolor[RGB]{255,205,205}18.9 & 48.8 & 40.1 & 54.9 \\
CMX~\cite{zhang2023cmx} &TITS'23  &MiT-B4  & 98.3 & 90.1 & 75.2 & 64.5 & 50.2 & 35.3 & 8.5 & 54.2 & \cellcolor[RGB]{255,205,205}60.6 & 59.7 \\
EAEFNet~\cite{liang2023explicit} & RAL'23 & ResNet-152  & 97.6 & 87.6 & 72.6 & 63.8 & 48.6 & 35.0 & 14.2 & 52.4 & 58.3 & 58.9 \\
CACFNet~\cite{zhou2023cacfnet} & TIV'23  & ConvNeXt-B  & 98.1 & 89.2 & 69.5 & 63.3 & 46.6 & 32.4 & 7.9 & 54.9 & 58.3 & 57.8 \\

SegMiF~\cite{liu2023multi} & ICCV'23  & MiT-B3  & 98.1 & 87.8 & 71.4 & 63.2 & 47.5 & 31.1 & 6.6 & 48.9 & 50.3 & 56.1 \\
CMNeXt~\cite{zhang2023delivering} & CVPR'23  & MiT-B4  & \cellcolor[RGB]{255,231,198}98.4 &\cellcolor[RGB]{255,205,205}91.5 & \cellcolor[RGB]{255,231,198}75.3 & \cellcolor[RGB]{255,231,198}67.6 & \cellcolor[RGB]{255,231,198}50.5 & 40.1 & 9.3 & 53.4 & 52.8 & 59.9 \\
CRM~\cite{shin2024complementary} & ICRA'24 & Swin-T & 98.2 & 90.0 & 73.1 & 63.7 & 47.9 & 40.7 & 9.9 & 54.4 & 54.2 & 59.1\\
MDBFNet~\cite{liang2024multi} & TIV'24 & ResNet-152 & 97.8 & 89.1 & 73.5 & 62.6 & 48.8 & 34.6 & 8.0 & 54.5 & 51.3 & 57.8 \\
Sigma~\cite{wan2025sigma} & WACV'25 & VMamba-T   & \cellcolor[RGB]{255,231,198}98.4 & 90.8 & 75.2 & 66.6 & 48.2 & 38.0 & 8.7 & 55.9 & \cellcolor[RGB]{255,231,198}60.4 & \cellcolor[RGB]{255,231,198}60.2 \\
HKDNet~\cite{tu2025hybrid} & TCSVT'25 & MiT-B3 & 98.0 & 87.1 & 71.7 & 63.1 & 42.9 & 38.1 & 0.3 & 55.5 & 51.8 & 56.5\\
OmniFuse~\cite{zhang2025omnifuse} & TPAMI'25 & - & \cellcolor[RGB]{255,205,205}98.6 & 87.3 & 69.7 & 63.9 & 37.7 & 33.7 & 13.5 & 53.8 & 48.5 & 56.3 \\
AGFNet~\cite{zhou2025agfnet} & TITS'25 & ResNet-50 & 98.2& 88.0 &73.7 & 64.6 & 45.3 & 34.3 & 13.8 & \cellcolor[RGB]{255,231,198}57.0 & 52.5 & 58.6\\
MixPrompt~\cite{haomixprompt} & NIPS'25 & MiT-B5 & 98.3 & 90.2 & 74.5 & 65.2 & 50.1 & \cellcolor[RGB]{255,205,205}48.3 & 10.5 & 51.7 & 52.0 & 60.1 \\
MiLNet~\cite{liu2025milnet} & TIP'25 & MiT-B3 & 98.2 & 87.9 & 72.3 & 64.5 & 46.2 & \cellcolor[RGB]{255,231,198}45.8 & 10.1 & 55.8 & 60.1 & 60.1 \\
\midrule
\textbf{RSGMamba-S} & \textbf{UR'26}  & \textbf{SegMAN-S} & \cellcolor[RGB]{255,231,198}\textbf{98.4} & \textbf{90.1} & \textbf{73.8} & \textbf{66.3} & \textbf{49.3} & \textbf{39.8} &\textbf{4.5} & \textbf{55.0} & \textbf{57.2} & \textbf{59.4}\\
\textbf{RSGMamba-B} & \textbf{UR'26}  & \textbf{SegMAN-B} & \cellcolor[RGB]{255,205,205}\textbf{98.6} &\cellcolor[RGB]{255,231,198}\textbf{91.0} &\cellcolor[RGB]{255,205,205}\textbf{75.5} & \cellcolor[RGB]{255,205,205}\textbf{67.9} & \cellcolor[RGB]{255,205,205}\textbf{50.6} & \textbf{41.3} & \textbf{8.6} & \cellcolor[RGB]{255,205,205}\textbf{57.2} & \textbf{59.4} & \cellcolor[RGB]{255,205,205}\textbf{61.1}\\
\bottomrule

\end{tabular}
}
\end{center}
\end{table*}

\begin{table}[!t]
\begin{center}
\caption{Quantitative comparisons on PST900 dataset. Bg: Background, HD: Hand-Drill, Bp: Backpack, FE: Fire-Extinguisher, and Sv: Survivor. The \colorbox[RGB]{255,205,205}{best} and \colorbox[RGB]{255,231,198}{second} best results are color-coded.}
\label{PST900}
\scalebox{0.9}{
\begin{tabular}{lcccccccc}
\toprule
\textbf{Method} & \textbf{Publication}    & \textbf{Bg} & \textbf{HD} & \textbf{Bp} & \textbf{FE} & \textbf{Sv} & \textbf{mIoU} \\
\midrule 
DSGBINet~\cite{xu2022dual} & TCSVT'22  & 99.4 & 75.0 & 85.1 & 79.3 & 75.6 & 82.9 \\
LASNet~\cite{li2022rgb} & TCSVT'23  & \cellcolor[RGB]{255,231,198}99.5 & 82.8 & 86.5 & 77.8 & 75.5 & 84.4 \\
FDCNet~\cite{zhao2022feature} & TCSVT'23 &99.2 & 70.4 & 72.2 & 71.5 & 72.4 & 77.1\\
SGFNet~\cite{wang2023sgfnet}{} & TCSVT'23 &99.4 & 76.7 & 85.4 &75.6 & 76.7 & 82.8\\
EAEFNet~\cite{liang2023explicit} & RAL'23  & \cellcolor[RGB]{255,231,198}99.5 & 83.9 & 87.7 & 80.4 &75.6 & 85.4\\
DPLNet~\cite{dong2024efficient} & IROS'24 & -& -& -& -& - & 86.7 \\
GoPT~\cite{xiong2024gopt} & AAAI'24 &-& -& -& -& - & 81.5 \\
CAINet~\cite{lv2024context} & TMM'24  & \cellcolor[RGB]{255,231,198}99.5 & 80.3 & 88.0 & 77.2 & 78.7 & 84.7\\
CRM~\cite{shin2024complementary} & ICRA'24 & \cellcolor[RGB]{255,231,198}99.5 & 79.1 & 86.0 & 86.2 & 78.7 & 85.9\\
MDBFNet~\cite{liang2024multi} & TIV'24 & \cellcolor[RGB]{255,205,205}99.6 & 76.1 & 88.7 & 84.4 & 80.7 &85.9 \\
CPAL~\cite{liu2025cpal} & TCSVT'25  & \cellcolor[RGB]{255,205,205}99.6 & \cellcolor[RGB]{255,205,205}84.9  & 87.9  & 81.2 & 80.4 & 86.8 \\
AGFNet~\cite{zhou2025agfnet} & TITS'25  & \cellcolor[RGB]{255,231,198}99.5 & 80.3 & 85.3 & \cellcolor[RGB]{255,231,198}90.2 & 78.7 & 84.8 \\
Sigma~\cite{wan2025sigma} & WACV'25 & \cellcolor[RGB]{255,205,205}99.6 & 81.9 & \cellcolor[RGB]{255,205,205}89.8 & 88.7 & \cellcolor[RGB]{255,205,205}82.7 & \cellcolor[RGB]{255,231,198}88.6 \\
MiLNet~\cite{liu2025milnet} & TIP'25 & \cellcolor[RGB]{255,231,198}99.5 & 76.3 & \cellcolor[RGB]{255,231,198}89.6 & 82.2 & 78.0 & 85.1\\
\midrule
\textbf{RSGMamba-S}  & \textbf{UR'26}  & \cellcolor[RGB]{255,231,198}\textbf{99.5} & \textbf{82.2} & \textbf{86.9} & \textbf{87.3} & \textbf{80.4} & \textbf{87.3}\\
\textbf{RSGMamba-B}  & \textbf{UR'26}  & \cellcolor[RGB]{255,205,205}\textbf{99.6} & \cellcolor[RGB]{255,231,198}\textbf{84.2} & \textbf{89.0} & \cellcolor[RGB]{255,205,205}\textbf{90.5} & \cellcolor[RGB]{255,231,198}\textbf{81.3} & \cellcolor[RGB]{255,205,205}\textbf{88.9}\\
\bottomrule

\end{tabular}
}
\end{center}
\end{table}

\subsection{Ablation Studies}

\textbf{Effectiveness of Key Components: RSGMB and LCGM.}
Table~\ref{tab:ablation_rsgmb_lem} investigates the individual and combined effects of RSGMB and LCGM on segmentation performance. As a baseline, the model without either module (ID 1) achieves an mIoU of 53.6\% with 24.31M parameters and 53.97G FLOPs. Introducing RSGMB alone (ID 2) yields a noticeable improvement of +1.3\% mIoU while incurring only a moderate increase in model complexity, indicating that reliability-aware state-space modeling effectively enhances cross-modal feature representation. Similarly, enabling LEM alone (ID 3) leads to a +1.7\% mIoU gain over the baseline, suggesting that explicit local enhancement plays a crucial role in refining spatial details and boundary-aware semantics. Although RSGMB introduces slightly higher computational overhead than LCGM, the resulting accuracy gain justifies its inclusion. When both RSGMB and LEM are jointly applied (ID 4), the model achieves the best performance, reaching 56.4\% mIoU and corresponding to a total improvement of +2.8\% mIoU over the baseline. This gain is notably larger than that obtained by either module alone, demonstrating their strong complementarity. Specifically, RSGMB focuses on reliable global cross-modal interaction, while LEM emphasizes local structural enhancement, together enabling more discriminative and robust feature learning. Overall, the results validate that the proposed design achieves a favorable trade-off between segmentation accuracy and computational cost. 

To investigate whether the projection layer in the fusion block should be parameterized as a full linear mapping or a lightweight low-rank adaptation, we replace the standard linear projection with a LoRA-based projection while keeping all other settings unchanged. Table~\ref{ablation_lora} shows that LoRA not only slightly improves segmentation accuracy from 56.2\% to 56.4\% mIoU (+0.2\%), but also reduces the model complexity, decreasing the number of parameters from 29.41M to 28.10M and FLOPs from 59.13G to 58.45G. These results suggest that the projection used in cross-modal fusion is highly redundant and can be effectively approximated by low-rank updates.

\begin{table}[!t]
\centering
\caption{Effectiveness of RSGMB and LCGM.}
\label{tab:ablation_rsgmb_lem}
\scalebox{1.0}{
\begin{tabular}{l|cccccc}
\toprule
\textbf{ID} & \textbf{RSGMB} & \textbf{LCGM} & \textbf{Params.}$\downarrow$ & \textbf{GFLOPs}$\downarrow$ & \textbf{mIoU}$\uparrow$ & $\boldsymbol{\Delta}$ \\
\midrule
1 & \XSolidBrush & \XSolidBrush & 24.31 & 53.97 & 53.6 & --\\
2 & \XSolidBrush & \Checkmark & 25.57 & 55.19 & 54.9 & \textcolor{seagreen}{+1.3} \\
3 & \Checkmark & \XSolidBrush & 26.83 & 57.22 & 55.3 & \textcolor{seagreen}{+1.7}\\
4 & \Checkmark & \Checkmark & 28.10 & 58.45 & 56.4 & \textcolor{seagreen}{+2.8}\\
\bottomrule
\end{tabular}}
\end{table}

\begin{table}[!t]
\begin{center}
\caption{Effectiveness of LORA projection.}
\label{ablation_lora}
\scalebox{1.0}{
\begin{tabular}{l|ccccc}
\toprule
\textbf{ID} & \textbf{Projection} & \textbf{Params.}$\downarrow$ & \textbf{GFLOPs}$\downarrow$ &\textbf{mIoU}$\uparrow$& \textbf{$\Delta$}\\
\midrule
1 & Linear & 29.41 & 59.13 & 56.2 &-\\
2 & LORA & 28.10  & 58.45 & 56.4 & \textcolor{seagreen}{+0.2}\\
\bottomrule
\end{tabular}
}
\end{center}
\end{table}

\begin{table}[t]
\centering
\caption{Effectiveness of the reliability modeling in RSGMB. \textbf{$g_u$} denotes the uncertainty-aware reliability estimation, and \textbf{$g_c$} denotes the cross-modal consistency modeling.}
\label{ablation:unccon}
\scalebox{1.0}{
\begin{tabular}{l|cccccc}
\toprule
\textbf{ID} & \textbf{Unc $g_u$} & \textbf{Con $g_c$} & \textbf{Params.}$\downarrow$ & \textbf{GFLOPs}$\downarrow$ &\textbf{mIoU}$\uparrow$& \textbf{$\Delta$}\\
\midrule
1 & \XSolidBrush & \XSolidBrush & 28.10 & 56.53 & 54.9 & -\\
2 & \Checkmark & \XSolidBrush &28.10 & 57.30 & 55.4 & \textcolor{seagreen}{+0.5}\\
3 & \XSolidBrush & \Checkmark &28.10 & 57.68 & 55.6 & \textcolor{seagreen}{+0.7}\\
4 & \Checkmark & \Checkmark &28.10 & 58.45 & 56.4 & \textcolor{seagreen}{+1.5}\\
\bottomrule
\end{tabular}
}
\end{table}

\begin{table}[t]
\centering
\caption{Comparisons of different fusion methods under the same backbone and decoder.}
\label{ablation:fusion}
\scalebox{1.0}{
\begin{tabular}{l|lcccc}
\toprule
\textbf{ID} &\textbf{Fusion} &  \textbf{Params.}$\downarrow$ & \textbf{GFLOPs}$\downarrow$ &\textbf{mIoU}$\uparrow$ & $\Delta$ \\
\midrule
1 &Concat  &30.96 & 60.83 & 54.7 & -\\
2 &Add  & 24.31 & 53.97 & 55.3 & \textcolor{seagreen}{+0.6}\\
3 &Cross-Attention  &35.42 & 135.06 & 54.0 & \textcolor{seagreen}{-0.5} \\
4 &Cross-Mamba  &26.83 &55.31 & 55.1 & \textcolor{seagreen}{+0.4}\\
5 &\textbf{RSGMB}  &26.83 & 57.22 & \textbf{56.4} & \textcolor{seagreen}{+1.7}\\
\bottomrule
\end{tabular}
}
\end{table}

\begin{figure}[!t]
\centering
\setlength{\tabcolsep}{0pt}
\renewcommand{\arraystretch}{1.0}

\begin{tabular}{@{}ccc@{}}

\ZoomCellSpyCircle{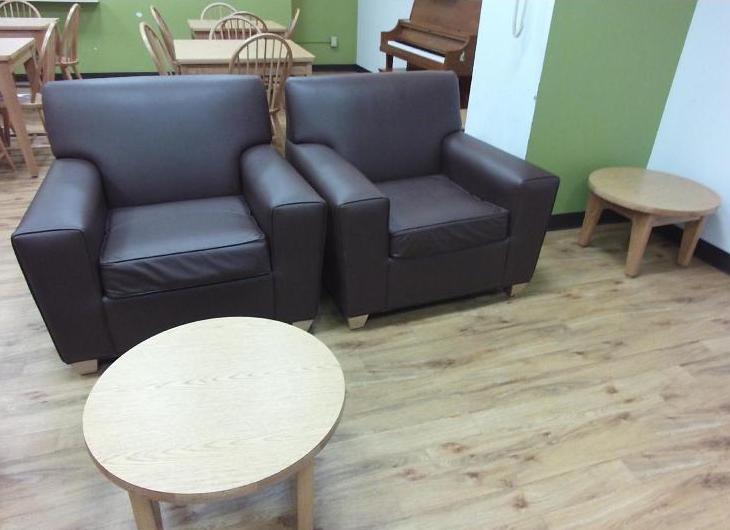}{0.162\textwidth}
  {\labA{MidLabel} {RGB img}}
  {cyan!60!black}{0.3}{0.52}  {1cm}{1.5cm}
&
\ZoomCellSpyCircle{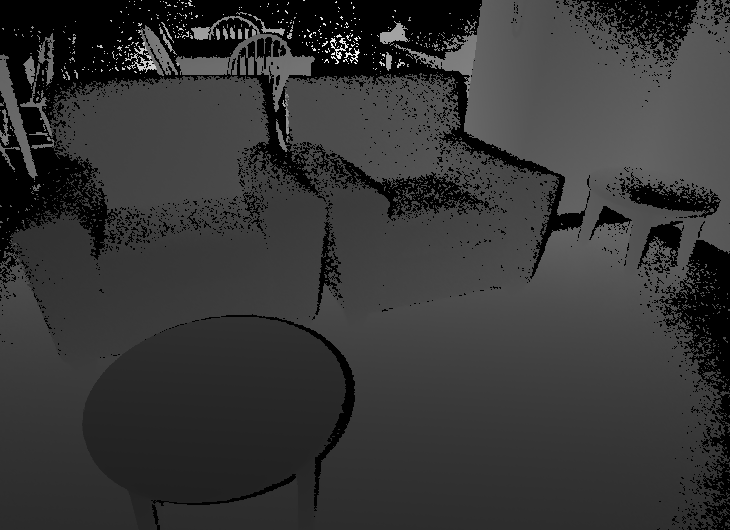}{0.16\textwidth}
  {\labA{MidLabel}{Depth img}}
  {orange}{0.3}{0.52}  {1cm}{1.5cm}
&
\ZoomCellSpyCircle{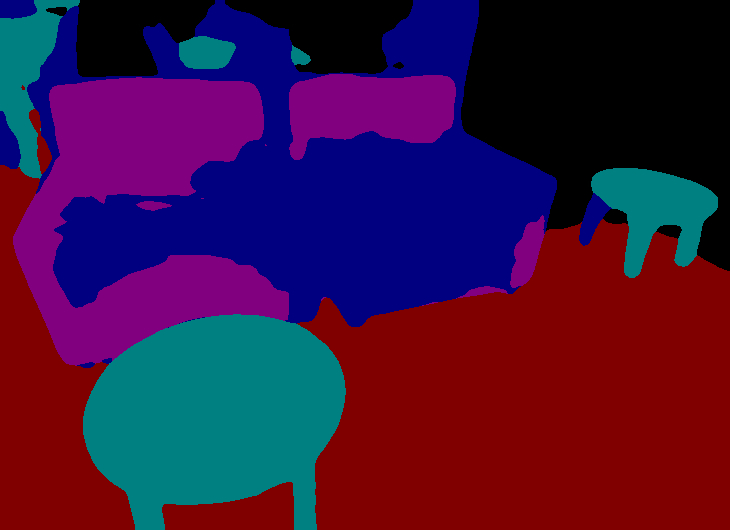}{0.16\textwidth}
  {\labA{MidLabel}{baseline}}
  {orange}{0.3}{0.52}  {1cm}{1.5cm}
\\
\ZoomCellSpyCircle{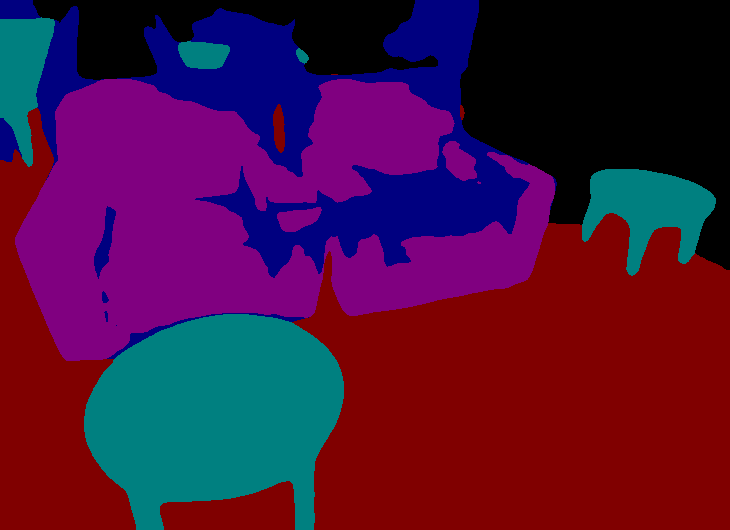}{0.16\textwidth}
  {\labA{MidLabel}{with \textit{\textbf{$g_u$}}}}
  {orange}{0.3}{0.52}  {1cm}{1.5cm}
&
\ZoomCellSpyCircle{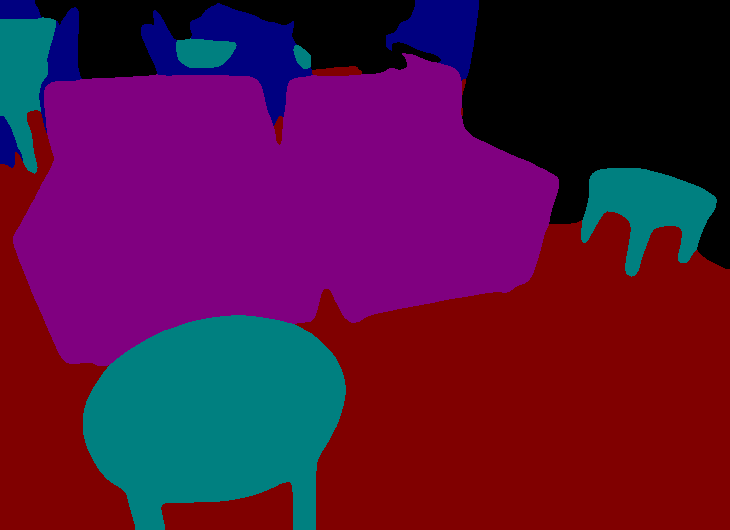}{0.16\textwidth}
  {\labA{MidLabel}{with \textit{\textbf{$g_u$}} \& \textit{\textbf{$g_c$}}}}
  {orange}{0.58}{0.78}  {1cm}{1.5cm}
&
\ZoomCellSpyCircle{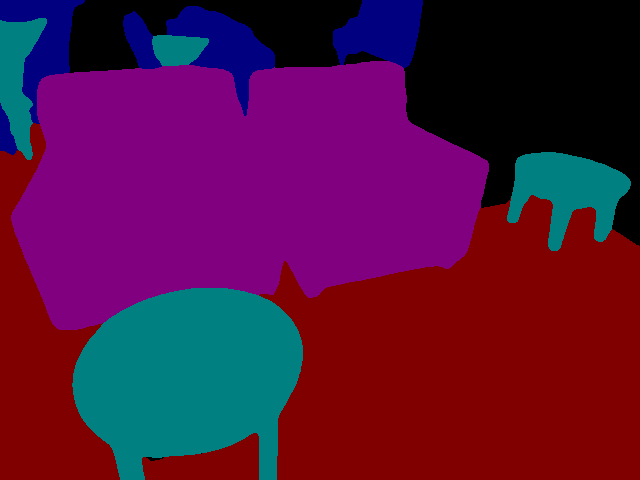}{0.156\textwidth}
  {\labA{MidLabel}{with \textit{RSGMB} \& \textit{LCGM}}}
  {orange}{0.58}{0.78}  {1cm}{1.5cm}
\end{tabular}
\caption{Qualitative comparisons of reliability modeling in our framework. The highlighted regions show that the uncertainty-aware module (\textit{\textbf{$g_u$}}) suppresses noisy depth-induced artifacts, while the consistency module (\textit{\textbf{$g_c$}}) enhances cross-modal semantic consistency. Their combination (\textit{\textbf{$g_u$}} \& \textit{\textbf{$g_c$}} = \textit{RSGMB}) produces more reliable and consistent predictions. Furthermore, with the integration of \textit{LCGM}, our final model achieves sharper boundaries and better preservation of fine structures.}
\label{fig:rsgb_lem}
\end{figure}

\textbf{Effectiveness of reliability modeling in RSGMB.}
To further verify the effectiveness of the proposed reliability modeling in RSGMB, we conduct a controlled ablation by progressively enabling two key components: uncertainty-aware reliability estimation ($g_u$) and cross-modal consistency modeling ($g_c$). As shown in Table~\ref{ablation:unccon}, starting from the variant without either component, introducing Unc leads to a clear improvement of +0.5\% mIoU, indicating that explicitly estimating modality reliability helps suppress noisy or ambiguous cross-modal cues. Enabling Con yields an even larger gain of +0.7\% mIoU, suggesting that enforcing cross-modal consistency provides a stronger supervisory signal for stabilizing fusion and reducing semantic ambiguity. When both components are jointly applied, the model achieves the best performance of 56.4\% mIoU, corresponding to a total gain of +1.5\% mIoU. As further illustrated by the qualitative results in Fig.~\ref{fig:rsgb_lem}, the baseline model suffers from noticeable misclassification in the sofa region, largely due to unreliable depth responses and weak cross-modal alignment. After incorporating Unc, the erroneous sofa predictions are visibly reduced, while Con further refines the segmentation by producing cleaner boundaries and more consistent semantic regions. These results demonstrate that uncertainty modeling and consistency modeling are complementary: the former improves the reliability of feature selection, while the latter strengthens cross-modal alignment, together enabling more robust and accurate RGB-X segmentation.

\textbf{Comparisons of different fusion methods.}
We conduct a comprehensive comparison of different fusion strategies on NYUDepth V2, using the same backbone (SegMAN-Small) for fair evaluation. The quantitative results are reported in Table~\ref{ablation:fusion}. We observe that simple Add achieves 55.3\% mIoU (+0.6\%) while reducing parameters and FLOPs compared to Concat, indicating that lightweight fusion can be sufficient without introducing severe cross-modal noise. In contrast, the standard Cross-Attention baseline ($Q$ from RGB, $K/V$ from depth) incurs the highest computational cost (135.06G FLOPs) but degrades performance (54.0\% mIoU, $-$0.5\%), suggesting that dense attention-based interaction is sensitive to depth noise, missing values, and misalignment in NYUDepth V2. As a result, unreliable cues may be propagated into RGB features, leading to feature contamination.

\begin{figure}[t]
\centering
\setlength{\tabcolsep}{0pt}
\renewcommand{\arraystretch}{1.0}

\begin{tabular}{@{}cccc@{}}

\ZoomCellSpyCircle{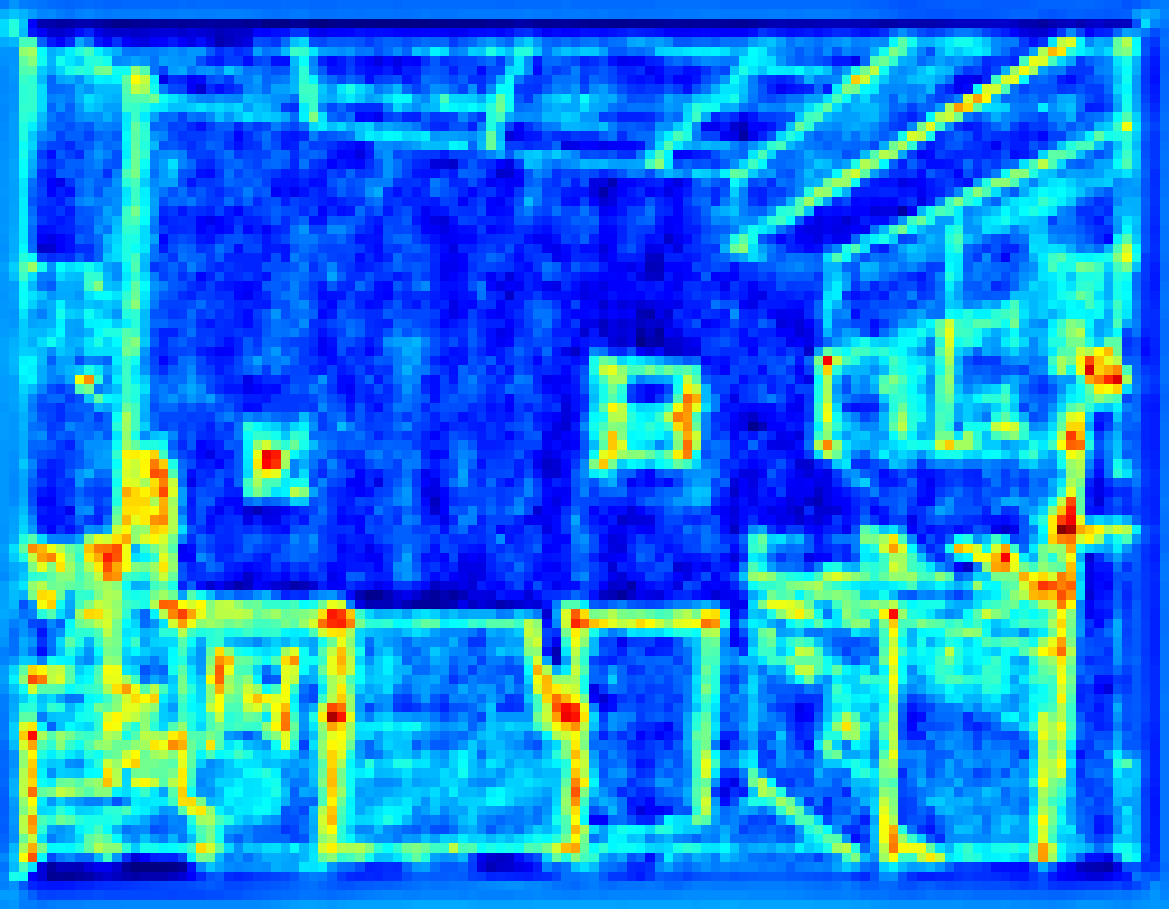}{0.12\textwidth}
  {}
  {orange}{0.7}{0.52}  {1cm}{1.5cm}

&\ZoomCellSpyCircle{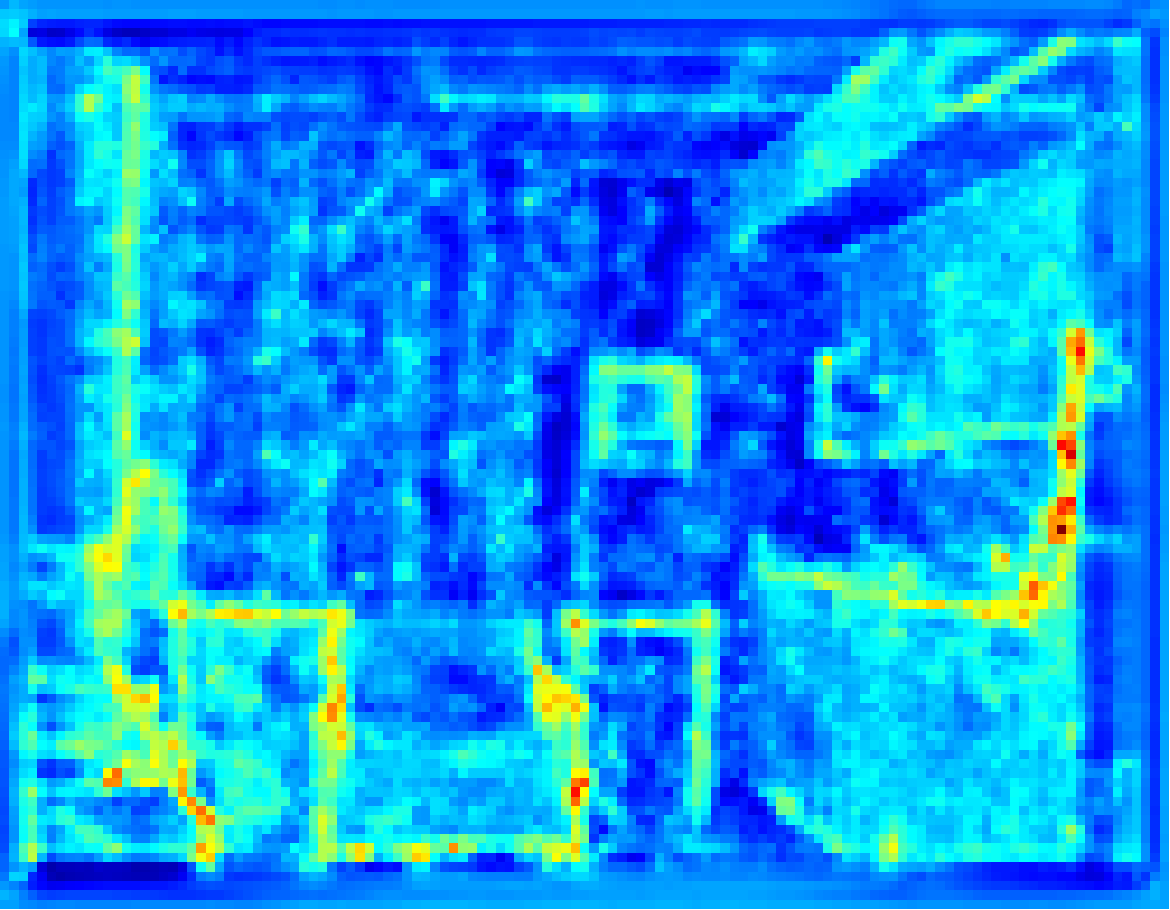}{0.12\textwidth}
  {}
  {orange}{0.7}{0.52}  {1cm}{1.5cm}
&
\ZoomCellSpyCircle{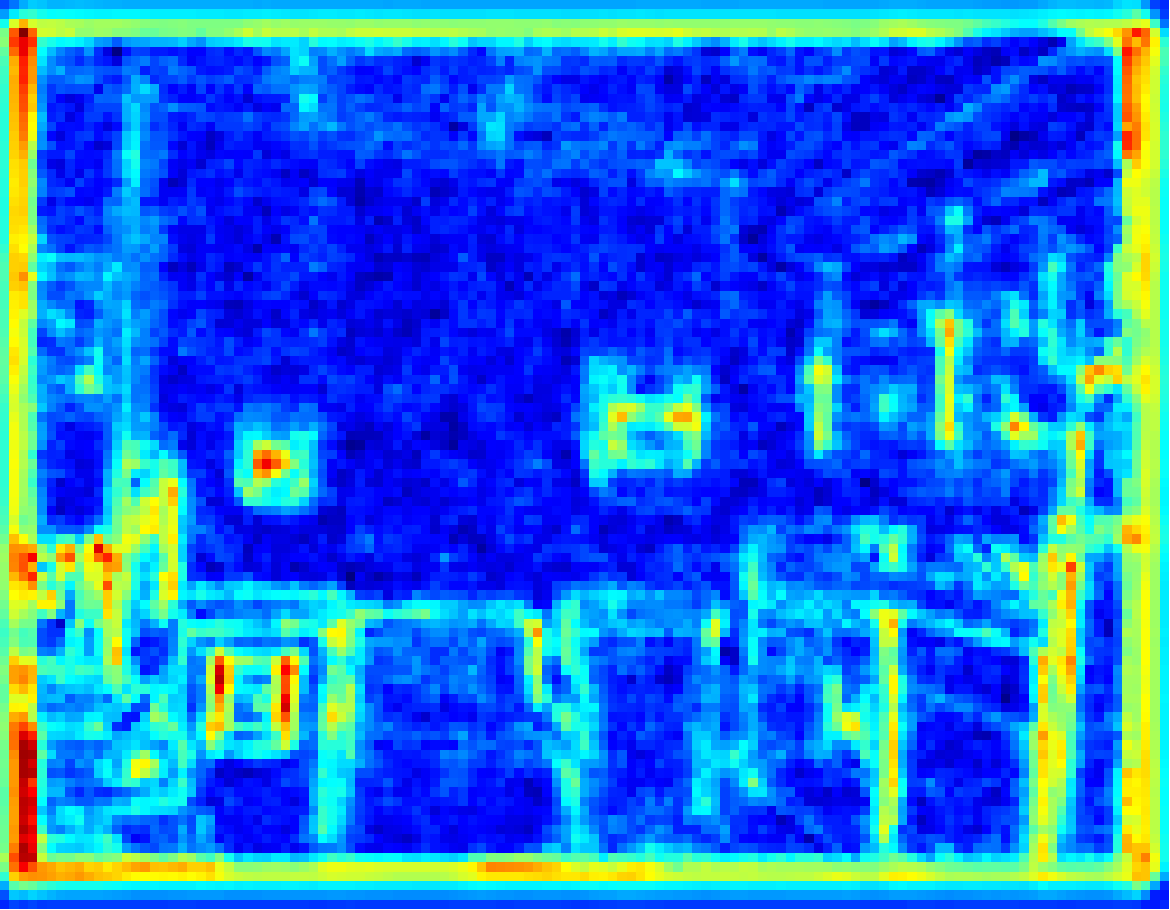}{0.12\textwidth}
  {}
  {orange}{0.7}{0.52}  {1cm}{1.5cm}
&
\ZoomCellSpyCircle{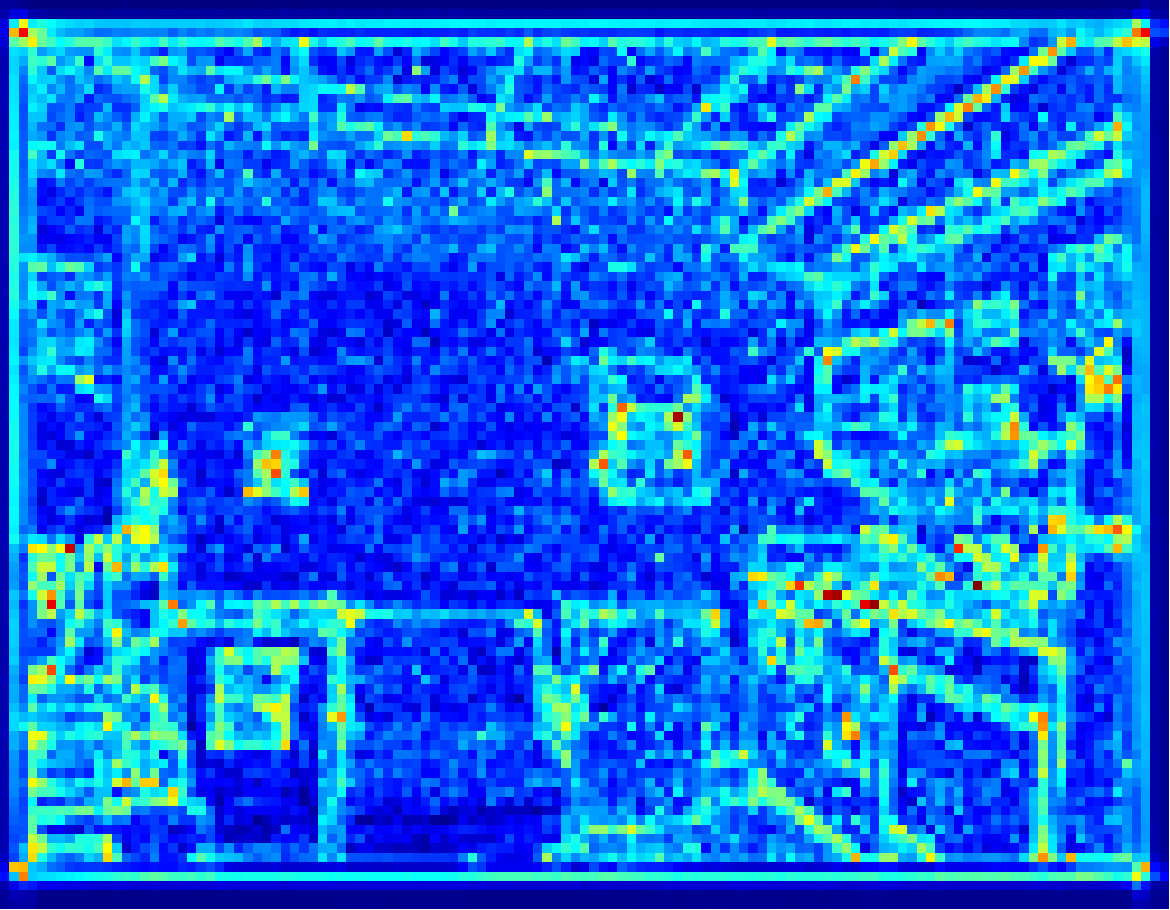}{0.12\textwidth}
  {}
  {orange}{0.7}{0.52}  {1cm}{1.5cm}
\\

\ZoomCellSpyCircle{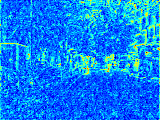}{0.12\textwidth}
  {\labA{MidLabel}{Addition}}
  {orange}{0.3}{0.25}  {1cm}{1.5cm}

&\ZoomCellSpyCircle{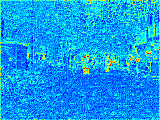}{0.12\textwidth}
  {\labA{MidLabel}{Cross-Attention}}
  {orange}{0.3}{0.25}  {1cm}{1.5cm}
&
\ZoomCellSpyCircle{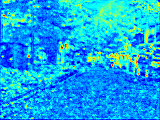}{0.12\textwidth}
  {\labA{MidLabel}{Cross-Mamba}}
  {orange}{0.3}{0.25}  {1cm}{1.5cm}
&
\ZoomCellSpyCircle{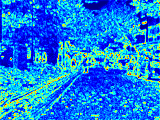}{0.12\textwidth}
  {\labA{OursLabel2}{\textbf{RSGMB (ours)}}}
  {orange}{0.3}{0.25}  {1cm}{1.5cm}

\end{tabular}
\caption{Visualization of feature responses under different fusion mechanisms: Addition, Cross-Attention, Cross-Mamba, and our proposed RSGMB. The zoomed-in regions indicate that our RSGMB produces clearer structural patterns and stronger activations in semantically meaningful regions while effectively suppressing noise, demonstrating its superior ability to model cross-modal interactions.}
\label{res:features}
\end{figure}

\begin{figure}[!t]
\centering
\setlength{\tabcolsep}{0pt}
\renewcommand{\arraystretch}{1.0}

\begin{tabular}{@{}cccc@{}}

\ZoomCellSpyCircle{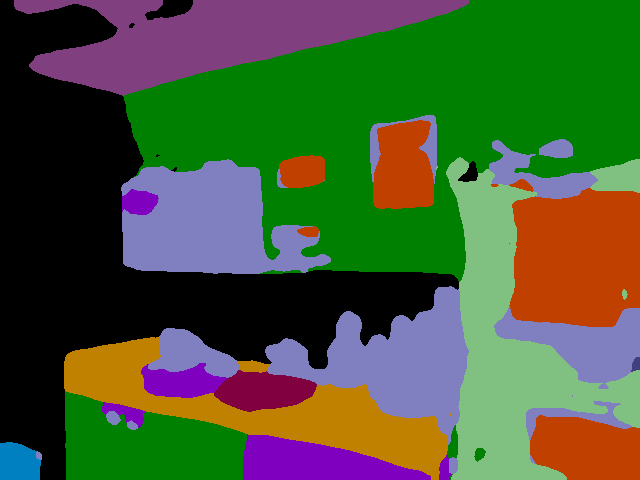}{0.12\textwidth}
  {}
  {orange}{0.3}{0.52}  {1cm}{1.5cm}

&\ZoomCellSpyCircle{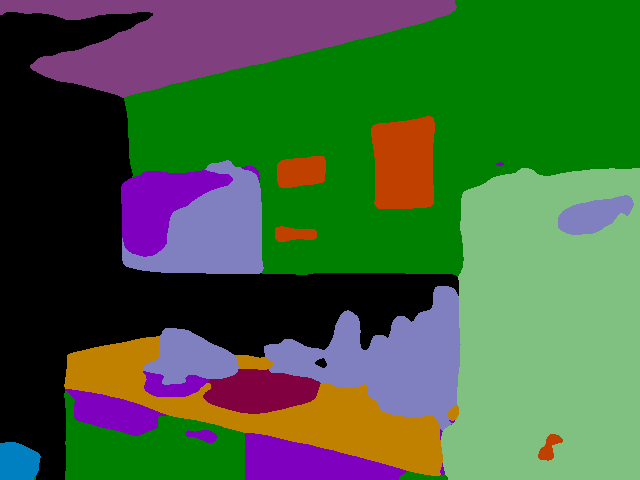}{0.12\textwidth}
  {}
  {orange}{0.3}{0.52}  {1cm}{1.5cm}
&
\ZoomCellSpyCircle{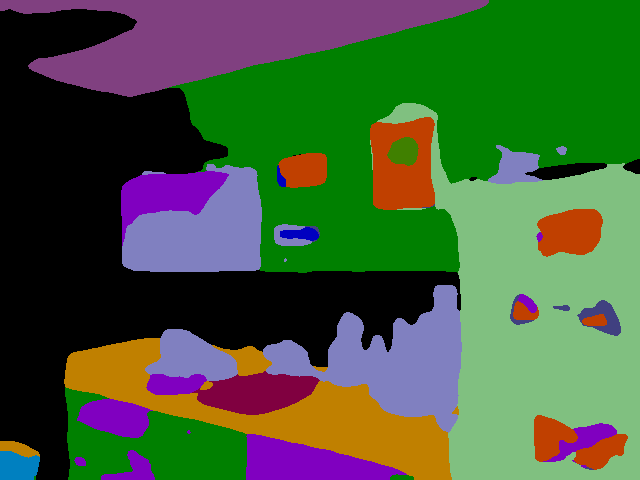}{0.12\textwidth}
  {}
  {orange}{0.3}{0.52}  {1cm}{1.5cm}
&
\ZoomCellSpyCircle{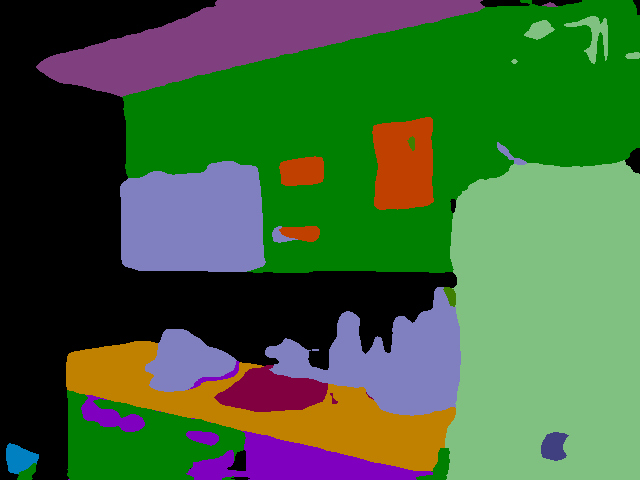}{0.12\textwidth}
  {}
  {orange}{0.3}{0.52}  {1cm}{1.5cm}
\\

\ZoomCellSpyCircle{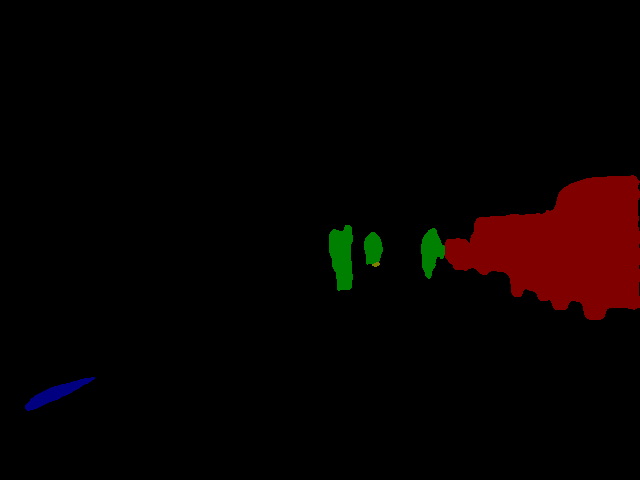}{0.12\textwidth}
  {\labA{MidLabel}{Addition}}
  {orange}{0.3}{0.25}  {1cm}{1.5cm}

&\ZoomCellSpyCircle{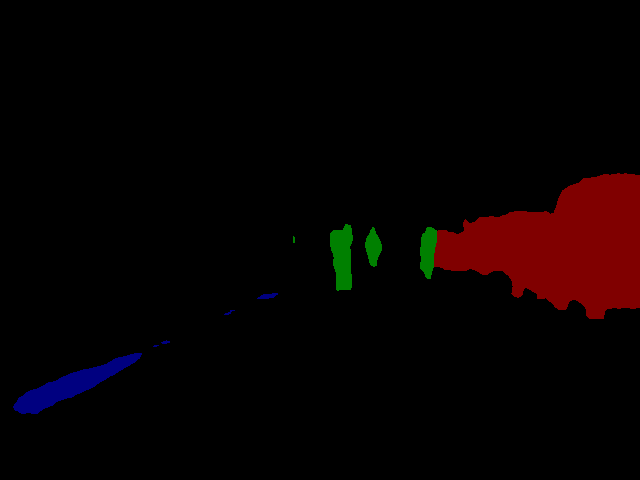}{0.12\textwidth}
  {\labA{MidLabel}{Cross-Attention}}
  {orange}{0.3}{0.25}  {1cm}{1.5cm}
&
\ZoomCellSpyCircle{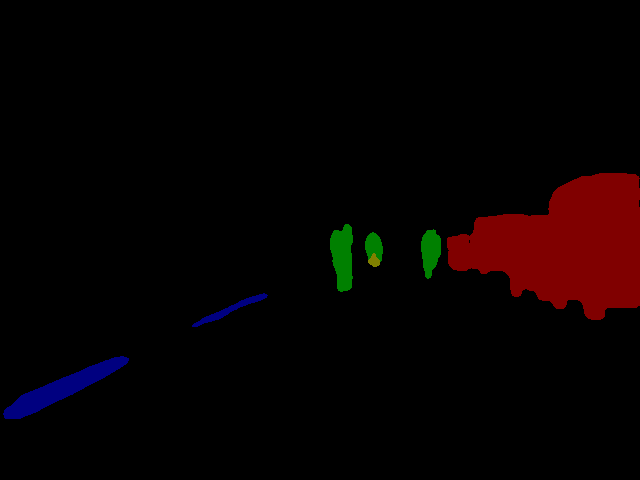}{0.12\textwidth}
  {\labA{MidLabel}{Cross-Mamba}}
  {orange}{0.3}{0.25}  {1cm}{1.5cm}
&
\ZoomCellSpyCircle{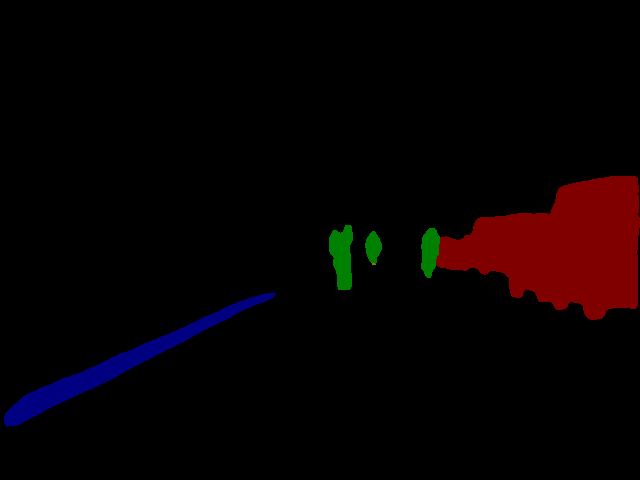}{0.12\textwidth}
  {\labA{OursLabel2}{\textbf{RSGMB (ours)}}}
  {orange}{0.3}{0.25}  {1cm}{1.5cm}

\end{tabular}
\caption{Visual comparisons of RGB-X semantic segmentation results. From left to right: RGB image, auxiliary modality, and segmentation predictions obtained by Addition, Cross-Attention, Cross-Mamba, and RSGMB (ours). The highlighted regions demonstrate that RSGMB better preserves structural details and thin objects while reducing noisy predictions.}
\label{res:comparison}
\end{figure}

Cross-Mamba provides a better accuracy–efficiency trade-off (55.1\% mIoU, +0.4\%) with much lower complexity, implying that SSM-based fusion is more stable than quadratic attention. Finally, RSGMB achieves the best performance (56.4\% mIoU, +1.7\%) with the same parameter budget as Cross-Mamba, demonstrating that the gain mainly comes from our reliability-aware design. By explicitly modeling uncertainty (Unc) and cross-modal consistency (Con), RSGMB selectively leverages informative depth cues while suppressing unreliable regions, leading to more robust fusion. As shown in Fig.~\ref{res:features}, existing fusion strategies such as Add and Cross-Attention tend to produce diffused or noisy activations, especially in regions where the auxiliary modality is unreliable. Although Cross-Mamba alleviates this issue to some extent, it still lacks the ability to explicitly distinguish reliable and unreliable cross-modal cues. In contrast, RSGMB generates more structured and concentrated feature responses, focusing on semantically meaningful regions while effectively suppressing noise, demonstrating the effectiveness of our reliability-aware gating mechanism in regulating cross-modal interactions.

This advantage further translates into improved segmentation quality. As illustrated in Fig.~\ref{res:comparison}, our method produces cleaner and more consistent predictions than other fusion strategies. In particular, RSGMB better preserves structural details and thin objects, while significantly reducing noisy and spurious predictions in challenging regions with depth ambiguity or missing values. These results confirm that explicitly modeling modality uncertainty and cross-modal consistency is critical for achieving robust and reliable RGB-X segmentation.

\section{Conclusion}
\label {sect:conclusion}
 In this paper, we presented RSGMamba, a reliability-aware fusion framework for multimodal semantic segmentation. Unlike prior methods that indiscriminately fuse heterogeneous modalities, RSGMamba explicitly models modality reliability and introduces a self-gated interaction mechanism to selectively exploit complementary cues while suppressing unreliable or noisy information. Built upon the linear-complexity formulation of state space models, RSGMamba achieves efficient global receptive field modeling and further enhances feature representations through a unified global–local enhancement strategy. Extensive experiments on both RGB-D and RGB-T benchmarks demonstrate consistent improvements over strong baselines, confirming the effectiveness, robustness, and efficiency of the proposed approach. More importantly, our results suggest that explicitly regulating cross-modal interaction according to modality reliability is crucial for robust multimodal segmentation.

\bibliographystyle{IEEEtran}
\bibliography{reference}

\end{document}